\documentclass{article} 
\usepackage[final]{colm2026_conference}

\usepackage{microtype}
\usepackage{hyperref}
\usepackage{url}
\usepackage{booktabs}

\usepackage{lineno}

\definecolor{darkblue}{rgb}{0, 0, 0.5}
\hypersetup{colorlinks=true, citecolor=darkblue, linkcolor=darkblue, urlcolor=darkblue}

\usepackage{graphicx}
\usepackage{tabularx}
\usepackage{array}
\usepackage{caption}
\usepackage{multirow}   
\usepackage{makecell}   
\usepackage{amsmath}    

\usepackage{listings}
\usepackage{color}
\usepackage{xcolor}  
\usepackage[most]{tcolorbox}
\usepackage{colortbl}

\definecolor{scaffoldblue}{RGB}{214,232,248}    
\definecolor{saturatepink}{RGB}{252,222,222}    
\definecolor{bestgreen}{RGB}{218,245,218}       
\definecolor{takeawaybg}{RGB}{224,242,238}      
\definecolor{takeawayborder}{RGB}{58,142,120}   
\definecolor{warnamber}{RGB}{255,243,217}       
\definecolor{emphgray}{RGB}{238,238,238}        
\definecolor{hlbestfg}{RGB}{14,100,44}          
\definecolor{hlsatfg}{RGB}{168,40,40}           
\definecolor{hlnumfg}{RGB}{30,80,155}           

\newtcolorbox{promptbox}[1][]{
  colback=scaffoldblue!15,
  colframe=scaffoldblue!60!black,
  fonttitle=\bfseries\small,
  boxrule=0.4pt,
  arc=1.5pt,
  left=2pt, right=2pt, top=2pt, bottom=2pt,
  breakable,
  title={#1}}
\newtcolorbox{findingbox}[1][]{
  colback=takeawaybg,
  colframe=takeawayborder,
  fonttitle=\bfseries\small,
  boxrule=0.4pt,
  arc=1.5pt,
  left=4pt, right=4pt, top=4pt, bottom=4pt,
  title={#1}}  
\newcommand{\hlbest}[1]{\textcolor{black}{\textbf{#1}}}
\newcommand{\hlsat}[1]{\textcolor{black}{#1}}
\newcommand{\hlnum}[1]{\textcolor{black}{#1}}

\definecolor{gray60}{gray}{0.4}
\definecolor{gray80}{gray}{0.25}

\DeclareTextCommand{\textquotedbl}{OT1}{\char34 }

\newcommand{\gbf}[1]{\textcolor{gray60}{\textbf{#1}}}
\newcommand{\gdbf}[1]{\textcolor{gray80}{\textbf{#1}}}
\newcolumntype{L}[1]{>{\raggedright\arraybackslash}p{#1}}
\newcolumntype{Y}{>{\raggedright\arraybackslash}X}

\title{Preserving What Matters: Semantic Scaffolds Beyond Saturation in Summarization Evaluation}

\author{
Ramin Fahimi\thanks{\ \ Equal contribution.} \And
Nikhil Reddy Pottanigari\footnotemark[1] \And
Noah Bolger \And
Sepideh Kharaghani \And
Ying Zhang \AND
\makebox[0.9\textwidth][c]{ServiceNow Canada} \\
\makebox[0.9\textwidth][c]{\texttt{\{ramin.fahimi, nikhilreddy.pottanigari,}} \\
\makebox[0.9\textwidth][c]{\texttt{noah.bolger,}} \\
\makebox[0.9\textwidth][c]{\texttt{sepideh.kharaghani, yin.zhang\}@servicenow.com}}
}

\begin{document}

\ifcolmsubmission
\linenumbers
\fi

\maketitle
\begin{abstract}

Summarization ships in countless production systems, making model selection a routine decision that depends on measuring summary quality. Existing metrics struggle to support this: ROUGE captures only surface overlap, while LLM-as-judge scores saturate to near-identical values that fail to rank models effectively. We observe this saturation across three public datasets, two proprietary datasets, and multilingual settings. Motivated by this, we introduce \textbf{Semantic Scaffold}, an evaluation framework that extracts a hierarchical representation of facts, questions, and entity attributes from a source text, labeling each as a main point or supporting detail, and reusing this structure as a fixed reference for scoring summaries. From this representation, we derive three diagnostic metrics: Fact Preservation Score (\textbf{FPS}), Question Preservation Score (\textbf{QPS}), and Entity Preservation Score (\textbf{EPS}), designed to reward the preservation of essential information while penalizing detail overload, and position them as interpretable diagnostics that remain informative where holistic axes collapse. Finally, we analyze four recurring failure modes of ROUGE and LLM-as-judge scores, demonstrating that scaffold-based evaluation remains informative where conventional metrics collapse.

\end{abstract}

\section{Introduction}
\label{sec:intro}

Document summarization, a long-standing NLP challenge~\citep{lin2004rouge,See2017GetTT}, is increasingly deployed in production settings, powering everything from search overviews to automated digests of reports. This ubiquity makes model selection a routine production decision that hinges on accurately measuring summary quality.

Existing evaluation methods fall short. ROUGE~\citep{lin2004rouge} and BLEU~\citep{papineni2002bleu} count n-gram overlap, missing semantic content in abstractive summaries. Human evaluation is reliable but costly at scale~\citep{fabbri2021summeval}. LLM-as-judge scoring~\citep{liu2023g,Zheng2023JudgingLW} is cheaper but poorly calibrated~\citep{fabbri2021summeval}, hard to interpret, and biased toward verbose text~\citep{Zheng2023JudgingLW}; moreover, automatic metrics in general struggle to reliably evaluate LLM-generated summaries~\citep{Goyal2022NewsSA,Zhang2023BenchmarkingLL}. On our internal Contracts dataset, the Hallucination axis spans only 0.00225 across eight models, while the widest holistic-axis span is 0.182 (Table~\ref{tab:results}). We call this loss of range near the top of the scale \emph{metric saturation}; compressed scores do not imply identical model quality.

\begin{figure}[!t]
    \centering
    \includegraphics[width=0.64\columnwidth]{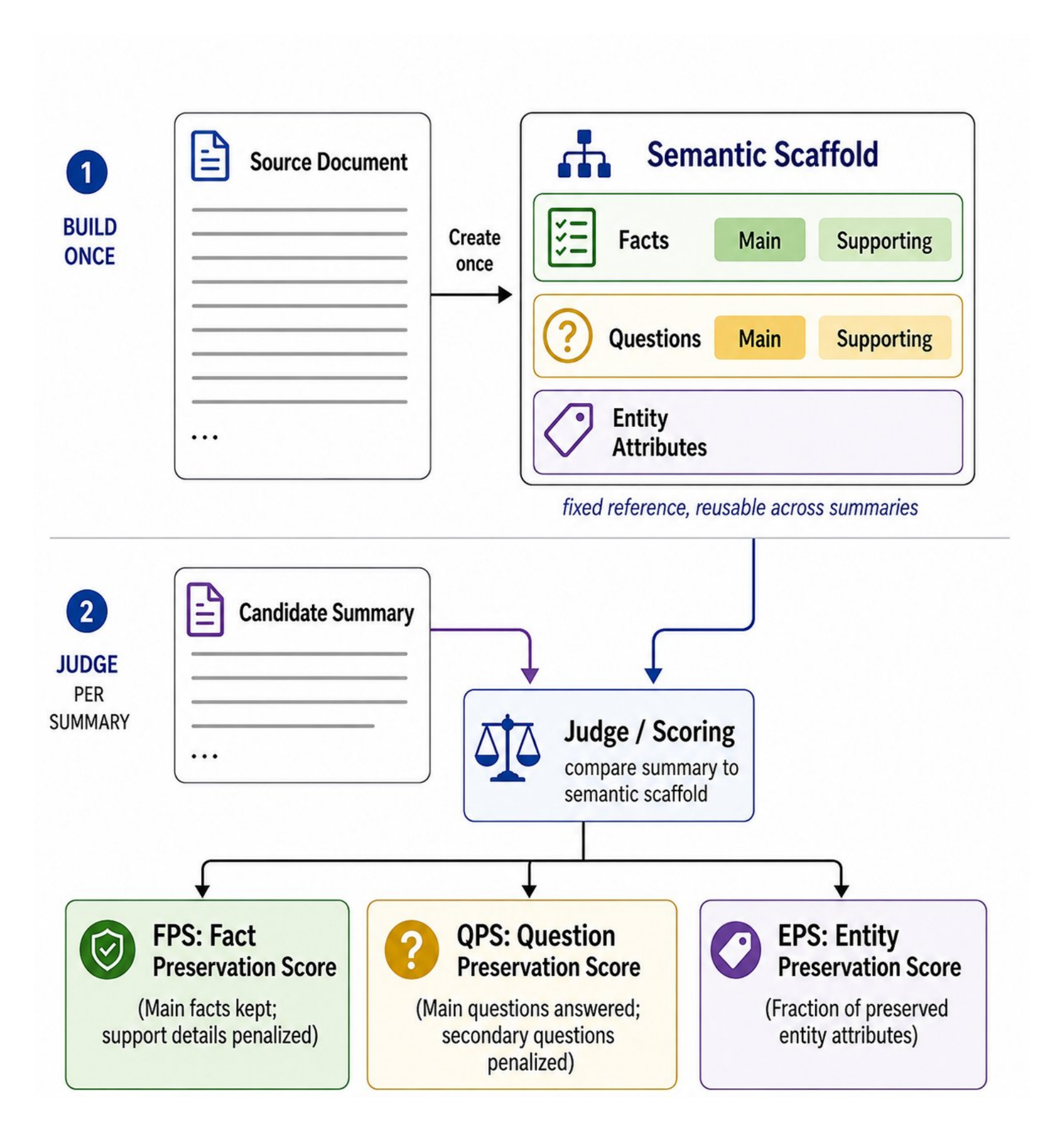}
    \caption{Overview of the Semantic Scaffold for summarization
    evaluation. }
    \label{fig:main}
\end{figure}

We propose the \textbf{Semantic Scaffold}, a proactive evaluation framework that addresses these limitations (Figure~\ref{fig:main}). The scaffold is a structured representation of a document's key facts, questions, hierarchy, and entities, extracted once from the source for reuse. Our decoupled ``extract-then-judge'' approach uses this hierarchy to distinguish between main facts/questions and supporting details. This enables a modular toolbox of interpretable metrics that specifically reward summaries for preserving the document's core structure by prioritizing main points over less critical details.

\noindent\textbf{Contributions.}\quad

(1)~We document substantial LLM-judge score compression across proprietary, public, and multilingual summarization settings;

(2)~we introduce the \textbf{Semantic Scaffold}, a reusable source representation that enables item-level evaluation of fact, question, and entity preservation; and

(3)~we catalogue four recurring failure modes of the ROUGE plus LLM-judge stack, each paired with the model-selection decision it corrupts.

\section{Related Work}
\label{sec:related}

\textbf{Reference-based.}
Lexical-overlap metrics, ROUGE~\citep{lin2004rouge},
BLEU~\citep{papineni2002bleu}, and METEOR~\citep{banerjee2005meteor},
rely on surface lexical matching and miss meaning and factual
accuracy~\citep{fabbri2021summeval}. Embedding metrics such as
BERTScore~\citep{zhang2019bertscore} and
MoverScore~\citep{zhao2019moverscore} capture deeper similarity but
still need references.

\textbf{Reference-free.}
Fact-based metrics and their meta-evaluations~\citep{gabriel2021go}, including SummaC~\citep{laban2022summac},
FactCC~\citep{kryscinski2020evaluating}, and
FineSurE~\citep{song2024finesure}, flag factual errors against the
source. QA-based metrics such as QAEval~\citep{deutsch2021towards} and
QuestEval~\citep{scialom2021questeval} generate questions and check
whether the summary answers them, but couple question generation to
scoring, so the questions cannot be reused. LLM-as-judge methods~\citep{Zheng2023JudgingLW} like
G-Eval~\citep{liu2023g} rate fluency, coherence, and factuality;
they track human judgments but remain poorly
calibrated~\citep{fabbri2021summeval} and
length-biased~\citep{Zheng2023JudgingLW}.

\textbf{From atomic units to structure.}
The Pyramid Method~\citep{nenkova-passonneau-2004-evaluating} scored summaries through
atomic content units; \citet{Liu2022RevisitingTG} formalized this as the
ACU protocol and AutoACU~\citep{Liu2023TowardsIA} automated it. Later
work adds factual verification and alignment:
FActScore~\citep{Min2023FActScoreFA}, ACUEval~\citep{wan2024acueval},
MiniCheck~\citep{Tang2024MiniCheckEF}, AlignScore~\citep{Zha2023AlignScoreEF},
and HAMLET~\citep{lee2025towards}; notably, FActScore finds frontier
LLMs below 60\% atomic factual precision in long-form generation.
The Semantic Scaffold extends this line with explicit
main/supporting hierarchy, a representation spanning facts, questions,
and entities, and a single extraction reused across models.

\section{Saturation: When LLM Judges Stop Ranking Models}
\label{sec:saturation}

Our goal was not to study metric saturation. However, when evaluating newer and cheaper models for production use cases, we discovered saturation in our LLM-as-a-judge scores, diminishing the usefulness of these metrics for selecting the top model. This section establishes this observation across proprietary production data, three public benchmarks and multilingual settings. 

\paragraph{Dataset.}
We use two annotated internal datasets and three public
datasets. \textbf{Screenshots}: 235 bug-report UI screenshots selected from
items with three available human references, with one reference retained per
instance (Figure~\ref{fig:screenshots}); short, structured, and entity-dense.
\textbf{Contracts}: 445 documents with dense legal language and
multi-thousand-word inputs, where
omissions and structural compression matter. We also evaluate on three public long-document benchmarks: \textbf{ArXiv-Sum}
\citep{Cohan2018ADA} (6,440 papers), \textbf{GovReport}
\citep{Huang2021EfficientAF} (973 reports), and \textbf{BigPatent}
\citep{Sharma2019BIGPATENTAL} (8,565 patents across nine CPC shards).

\paragraph{Models.}
We evaluate eight frontier models on the proprietary datasets:
GPT-4o-mini, GPT-4.1-mini, GPT-4o, GPT-4.1,
Gemini-2.0-Flash, Gemini-2.5-Flash,
Gemini-2.5-Pro, and Claude-3.7-Sonnet.
For the \textbf{Screenshots} dataset, we additionally include a \emph{Human} reference row as a target anchor.

We denote the three proprietary systems as \texttt{Internal-S},
\texttt{Internal-M}, and \texttt{Internal-L}. For public benchmark
evaluation, we use five models: \texttt{Internal-S}, \texttt{Internal-L},
GPT-4.1-mini, Gemini-2.5-Flash, and Claude-3.7-Sonnet; the qualitative
walkthrough in Appendix~\ref{app:walkthrough} also includes \texttt{Internal-M}.

\paragraph{Observed saturation.}
We evaluate every summary with a G-Eval-style~\citep{liu2023g} LLM judge on four holistic axes (Faithfulness, Completeness, Understandability, and Hallucination; prompt in \S\ref{app:direct_score}).

\begin{table}[t]
\centering\small
\setlength{\tabcolsep}{6pt}
\begin{tabular}{p{3cm}cc}
\toprule
Dataset \footnotemark[1] & DS range & Hall.\ range \\
\midrule
ArXiv-Sum   & \cellcolor{saturatepink!20}9.01 to 9.42 & \cellcolor{saturatepink}9.99 to 10.00 \\
GovReport  & \cellcolor{saturatepink!20}9.00 to 9.34 & \cellcolor{saturatepink}9.97 to 10.00 \\
BigPatent  & \cellcolor{saturatepink!20}9.23 to 9.55 & \cellcolor{saturatepink}10.00 to 10.00 \\
\bottomrule
\end{tabular}
\caption{Direct-scoring (DS) saturation on three public benchmarks, five
models each. The largest unrounded DS spread is 0.412; every Hallucination
mean is at least 9.972.}
\label{tab:saturation}
\end{table}
\footnotetext[1]{ArXiv-Sum \citep{Cohan2018ADA}, GovReport \citep{Huang2021EfficientAF}, BigPatent \citep{Sharma2019BIGPATENTAL}.}

\textbf{Proprietary datasets.}
On Contracts, every Hallucination mean rounds to \hlsat{1.00}; the
unrounded span is 0.00225. Faithfulness spans 0.057 and Understandability
0.025. Screenshots shows similar compression: Hallucination spans 0.025 and
Understandability 0.040. Per-model scores are in Table~\ref{tab:results}.

\textbf{Public benchmarks.}
Saturation is not confined to our internal data. On three widely used public
benchmarks scored on a 1--10 scale, the widest DS spread across five models
is \hlsat{0.412}, and Hallucination is at least \hlsat{9.972} in every cell
(Table~\ref{tab:saturation}).

\begin{table}[!h]
\centering\small
\setlength{\tabcolsep}{4pt}
\begin{tabular}{llcc}
\toprule
Model & Lang.\ group & Faith.\ avg & Hall.\ avg \\
\midrule
\multirow{2}{*}{Gemini-2.5-Flash}
 & English     & 9.67 & \cellcolor{saturatepink!25}9.99 \\
 & Non-English & 9.63 & \cellcolor{saturatepink!25}9.99 \\
\addlinespace
\multirow{2}{*}{GPT-4.1-mini}
 & English     & 9.57 & \cellcolor{saturatepink!25}9.97 \\
 & Non-English & 9.55 & \cellcolor{saturatepink!25}9.97 \\
\addlinespace
\multirow{2}{*}{Claude-3.7-Sonnet}
 & English     & 9.73 & \cellcolor{saturatepink!25}9.99 \\
 & Non-English & 9.66 & \cellcolor{saturatepink!25}9.98 \\
\bottomrule
\end{tabular}
\caption{Cross-lingual saturation on three internal screenshot datasets
(Faith./Hall.\ averaged over English vs.\ nine other languages).}
\label{tab:multilingual}
\end{table}

\textbf{Multilingual evidence.}
Across three models, three datasets, and ten languages (90 aggregate cells),
Hallucination ranges from 9.95 to 10.00
(Table~\ref{tab:multilingual}). OCR text and judge prompts were professionally
translated into nine languages with human review. These data show
compression in this translated pipeline, not a causal effect of language.
Per-language breakdowns are in Appendix~\ref{app:crosslingual}.

\section{Semantic Scaffold}
Saturation is a measurement failure, the score stops discriminating exactly when generation quality and decision stakes are highest. Motivated by this, we introduce the Semantic Scaffold.
\label{sec:method}

\subsection{Architecture}
\label{ssec:architecture}

The scaffold has three properties: It is \textbf{proactive}, with evaluation targets fixed before seeing any summary to prevent criteria drift; \textbf{decoupled}, as extraction runs once while judgment runs per candidate; and \textbf{reusable}, with one scaffold serving all summaries of a source for consistent, cheap comparisons.

\paragraph{Two stages.}
\emph{Creation} uses one LLM pass to extract a structured representation of facts, questions, hierarchy, and entities from the source.
\emph{Judgment}  then scores each candidate against this fixed reference. Each score links to a scaffold item, providing item-level evidence of what a summary kept, dropped, or misrepresented, rather than a single holistic verdict.

\paragraph{Components.}
Each scaffold item is tagged a \emph{main point} or a \emph{supporting
detail}. \textbf{Atomic facts} are concise, self-contained statements
(20--100 per document). \textbf{Key questions} capture what the
document answers, each with a brief expected answer and source-sentence
provenance. \textbf{Entity attributes} list salient entities (people,
organizations, products, concepts) and their defining properties for
entity-level checks. Templates are in Appendix~\ref{app:prompts}.

\subsection{Baseline Metrics}
\label{ssec:toolbox}

Our method builds on FineSurE~\citep{song2024finesure}, adopting its two-stage structure and its \textbf{Fact Faithfulness (F)} and \textbf{Fact Coverage (FC)} metrics. We add \textbf{Question Coverage (QC)} as a third baseline (Table~\ref{tab:foundational}). However, FineSurE lacks hierarchy, treating all facts equally. Our scaffold metrics extend it by tagging all facts and questions as main or supporting and rewarding summaries that prioritize the main tier. Appendix~\ref{app:metrics_summary} summarizes all metric definitions; formulas for FPS, QPS, and EPS follow below.

\begin{table}[t]
\centering\small
\setlength{\tabcolsep}{3pt}
\begin{tabular}{@{}lp{4.8cm}p{4cm}@{}}
\toprule
\textbf{Metric} & \textbf{Formula} & \textbf{What it measures} \\
\midrule
F  & faithful / total facts &
Fact faithfulness \footnotemark[2]  \\

FC & covered / total facts  &
Fact coverage \footnotemark[2]  \\

QC & answered / total questions &
Question coverage \\
\bottomrule
\end{tabular}
\caption{Baseline evaluation metrics.}
\label{tab:foundational}
\end{table}

\footnotetext[2]{Built on FineSurE \citep{song2024finesure}.}

\subsection{Scaffold Metrics}
\label{ssec:showcase}
\paragraph{Fact Preservation Score (FPS).}
FPS rewards keeping main facts and penalizes supporting ones. 
\begin{equation}
\mathrm{FPS}=\frac{M_s}{M_c}\times\left(1-\frac{S_s}{S_c}\right)
\label{eq:fps}
\end{equation}
where $M_s$ and $S_s$ denote main and supporting facts preserved in the summary, 
and $M_c$ and $S_c$ denote main and supporting facts in the scaffold. The first term measures the share of main facts preserved; the second term, ($1-S_s/S_c$), penalizes inclusion of supporting facts. Reproducing every detail is closer to paraphrase than summary; stating all supporting facts ($S_s=S_c$) drives FPS to $0$ regardless of main-fact coverage.

\paragraph{Question Preservation Score (QPS).}
QPS is the question analogue, rewarding answers to main questions and
penalizing attention to secondary ones.
\begin{equation}
\mathrm{QPS}=\frac{Q_s}{Q_c}\times\left(1-\frac{R_s}{R_c}\right)
\end{equation}
with $Q_s$ and $R_s$ denoting main and supporting questions answered in the summary,
and $Q_c$ and $R_c$ denoting main and supporting questions in the scaffold.

\paragraph{Entity Preservation Score (EPS).}
Entity fidelity is especially important in domains such as legal, medical, and financial text,
where a dropped attribute changes meaning. Unlike FineSurE's binary
entity check, EPS is continuous, the fraction of scaffold entity
attributes preserved:
\begin{equation}
\mathrm{EPS}=\frac{|\text{preserved attributes}|}{|\text{scaffold attributes}|}
\end{equation}
so it distinguishes summaries that keep some attributes from those that
keep none.

\begin{findingbox}[Key Point: Summary goals drive metric choice]
The three scaffold metrics are not interchangeable; each metric supports a different summary goal. \emph{\hlnum{EPS} leads when entity context must be preserved} (e.g. contracts, financial/medical text, identifier-dense screenshots). \emph{\hlnum{QPS} leads when the summary must answer main questions} (e.g. triage notes, executive digests). \emph{\hlnum{FPS} leads when a compressed abstract summary is required} (e.g. briefings, news-style digests).
\end{findingbox}

\paragraph{Implementation.}

We use GPT-4.1 for both extraction and judgment (temperature $0.1$, \texttt{max\_tokens}\,=\,4096) and score models zero-shot in their default configurations. A single model builds the scaffold and judges against it so that every candidate is scored against an identical reference; this is what makes the scaffold reusable across models and comparisons cheap.

\section{Main Results}
\label{sec:experiments}

Table~\ref{tab:results} reports all metrics across two proprietary
datasets, Contracts and Screenshots.

\begin{table*}[t]
\centering\small
\setlength{\tabcolsep}{4pt}
\resizebox{\textwidth}{!}{%
\begin{tabular}{llcccccccccc}
\toprule
Dataset & Model & \cellcolor{saturatepink!40}Faith. & \cellcolor{saturatepink!40}Compl. & \cellcolor{saturatepink!40}Underst. & \cellcolor{saturatepink!40}Hall. &
  F$^\dagger$ & FC$^\dagger$ & \cellcolor{scaffoldblue!50}FPS$^\ddagger$ & QC & \cellcolor{scaffoldblue!50}QPS$^\ddagger$ & \cellcolor{scaffoldblue!50}EPS \\
\midrule
\multirow{9}{*}{\makecell[l]{Screenshots}}
& GPT-4o-mini      & 0.89 & 0.82 & 0.96 & 0.97 & 0.93 & 0.43 & 0.51 & 0.36 & \cellcolor{bestgreen}\textbf{0.73} & 0.79 \\
& GPT-4.1-mini     & 0.93 & 0.87 & 0.97 & 0.99 & 0.97 & 0.49 & \gdbf{0.52} & 0.42 & \gdbf{0.70} & \gdbf{0.87} \\
& Gemini-2.0-Flash & 0.96 & 0.86 & 0.98 & 1.00 & 0.98 & 0.48 & 0.51 & 0.42 & 0.69 & 0.85 \\
& Gemini-2.5-Flash & 0.95 & 0.87 & 0.97 & 1.00 & \cellcolor{bestgreen}\gbf{0.99} & 0.53 & \cellcolor{bestgreen}\textbf{0.53} & 0.45 & 0.67 & \gbf{0.88} \\
& GPT-4o           & 0.87 & 0.77 & 0.95 & 0.98 & 0.97 & 0.31 & 0.39 & 0.25 & 0.70 & 0.71 \\
& GPT-4.1          & 0.92 & 0.85 & 0.97 & 0.99 & 0.97 & 0.47 & \gdbf{0.52} & 0.39 & \gdbf{0.71} & \gdbf{0.86} \\
& Gemini-2.5-Pro   & 0.95 & 0.89 & 0.98 & 0.99 & 0.97 & 0.55 & 0.50 & 0.47 & 0.65 & 0.85 \\
& Claude-3.7-Sonnet& \cellcolor{bestgreen}\gbf{0.97} & \cellcolor{bestgreen}\gbf{0.90} & \cellcolor{bestgreen}\gbf{0.99} & \gbf{1.00} & 0.98 & \cellcolor{bestgreen}\gbf{0.57} & 0.47 & \cellcolor{bestgreen}\gbf{0.50} & 0.62 & \cellcolor{bestgreen}\textbf{0.90} \\\cmidrule(l){2-12}
& \cellcolor{emphgray}\textit{Human}    & \cellcolor{emphgray}0.95 & \cellcolor{emphgray}0.86 & \cellcolor{emphgray}0.98 & \cellcolor{emphgray}0.99 & \cellcolor{emphgray}0.91 & \cellcolor{emphgray}0.53 & \cellcolor{emphgray}0.40 & \cellcolor{emphgray}0.44 & \cellcolor{emphgray}0.60 & \cellcolor{emphgray}0.76 \\
\midrule
\multirow{8}{*}{\makecell[l]{Contracts}}
& GPT-4o-mini      & 0.92 & 0.81 & 0.96 & \cellcolor{saturatepink}1.00 & 0.98 & 0.14 & 0.39 & 0.19 & 0.60 & 0.28 \\
& GPT-4.1-mini     & 0.97 & \cellcolor{bestgreen}\gbf{0.89} & 0.96 & \cellcolor{saturatepink}1.00 & 0.98 & \cellcolor{bestgreen}\gbf{0.20} & \cellcolor{bestgreen}\textbf{0.50} & \cellcolor{bestgreen}\gbf{0.28} & \cellcolor{bestgreen}\textbf{0.75} & \cellcolor{bestgreen}\textbf{0.46} \\
& Gemini-2.0-Flash & 0.95 & 0.82 & 0.96 & \cellcolor{saturatepink}1.00 & 0.98 & 0.15 & 0.41 & 0.21 & 0.61 & 0.30 \\
& Gemini-2.5-Flash & 0.95 & 0.84 & 0.97 & \cellcolor{saturatepink}1.00 & 0.95 & 0.15 & 0.42 & 0.21 & 0.63 & 0.31 \\
& GPT-4o           & 0.95 & 0.71 & 0.96 & \cellcolor{saturatepink}1.00 & \cellcolor{bestgreen}\gbf{1.00} & 0.06 & 0.16 & 0.09 & 0.27 & 0.14 \\
& GPT-4.1          & \cellcolor{bestgreen}\gbf{0.98} & 0.86 & 0.97 & \cellcolor{saturatepink}1.00 & 0.99 & 0.15 & 0.43 & 0.22 & 0.64 & 0.38 \\
& Gemini-2.5-Pro   & 0.93 & 0.83 & 0.98 & \cellcolor{saturatepink}1.00 & 0.95 & 0.15 & 0.42 & 0.21 & 0.63 & 0.31 \\
& Claude-3.7-Sonnet& 0.97 & 0.87 & \cellcolor{bestgreen}\gbf{0.98} & \cellcolor{saturatepink}1.00 & 0.96 & 0.17 & 0.46 & 0.24 & 0.67 & 0.36 \\
\bottomrule
\end{tabular}
}
\caption{Descriptive scores on both proprietary datasets. On Contracts,
Hallucination spans 0.002, Understandability 0.025, Faithfulness 0.057,
and Completeness 0.182; EPS spans 0.32.
The strongest value per column is highlighted in \colorbox{bestgreen}{\strut green}; \colorbox{saturatepink}{\strut pink} marks saturated cells. The \colorbox{emphgray}{\strut Human}
row (Screenshots only) is an anchor, not a ceiling.
\noindent$^\dagger$\citet{song2024finesure}}
\label{tab:results}
\smallskip
\end{table*}

\paragraph{Scaffold metrics restore discrimination.}

\begin{figure}[!t]
\centering
\includegraphics[width=0.54\columnwidth]{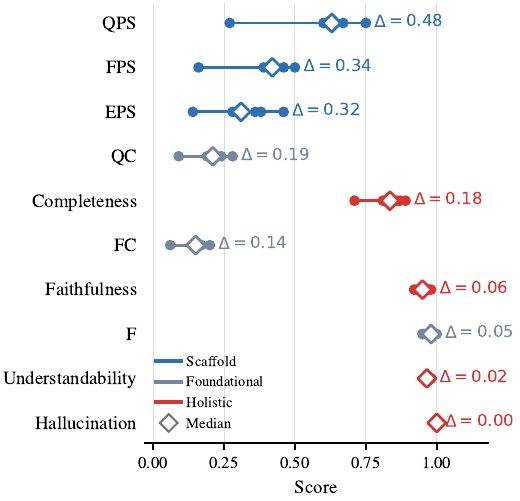}
\caption{Score range across 8 models on Contracts, sorted by spread (dots = models). Holistic metrics
(red) saturate near 1.0; scaffold metrics (blue) retain spread;
foundational (gray) in between.}
\label{fig:dynamic_range}
\end{figure}

Where holistic metrics saturate (~\S\ref{sec:saturation}), scaffold metrics maintain separation: QPS, FPS, and EPS span \hlbest{$\Delta = 0.32$ to $0.48$} on Contracts, compared to Hallucination (\hlsat{$\Delta = 0.00$}), Understandability (\hlsat{$0.02$}), and Faithfulness (\hlsat{$0.06$}) (Figure~\ref{fig:dynamic_range}). This represents a wider spread, making model differences visible where holistic axes collapse. Even when baseline metrics like FC and QC cannot differentiate (e.g., 0.14 and 0.19 spreads), 

scaffold metrics retain resolution, keeping model differences visible where holistic axes no longer distinguish systems. We report spread as resolving power; pairing it with a preference target is the natural next step for using these axes to select models.

\paragraph{Themes are easy; hierarchy is hard.}
A systematic gap between QPS and FPS scores reveals a core limitation in current LLMs. Across every model and dataset, QPS exceeds FPS (e.g., GPT-4.1-mini on Contracts: QPS $= 0.75$ vs.\ FPS $= 0.50$). The two criteria are calibrated differently by design: question alignment credits a question whenever the summary supplies enough information to answer it, while fact alignment requires the fact itself to be conveyed (\S\ref{app:qalign}). The gap therefore locates where a summary's coverage sits between thematic and propositional---models track a document's agenda more readily than its fact hierarchy---and raising FPS at fixed QPS and EPS is a well-posed target under matched criteria.

\begin{figure}[!t]
\centering
\includegraphics[width=0.58\columnwidth]{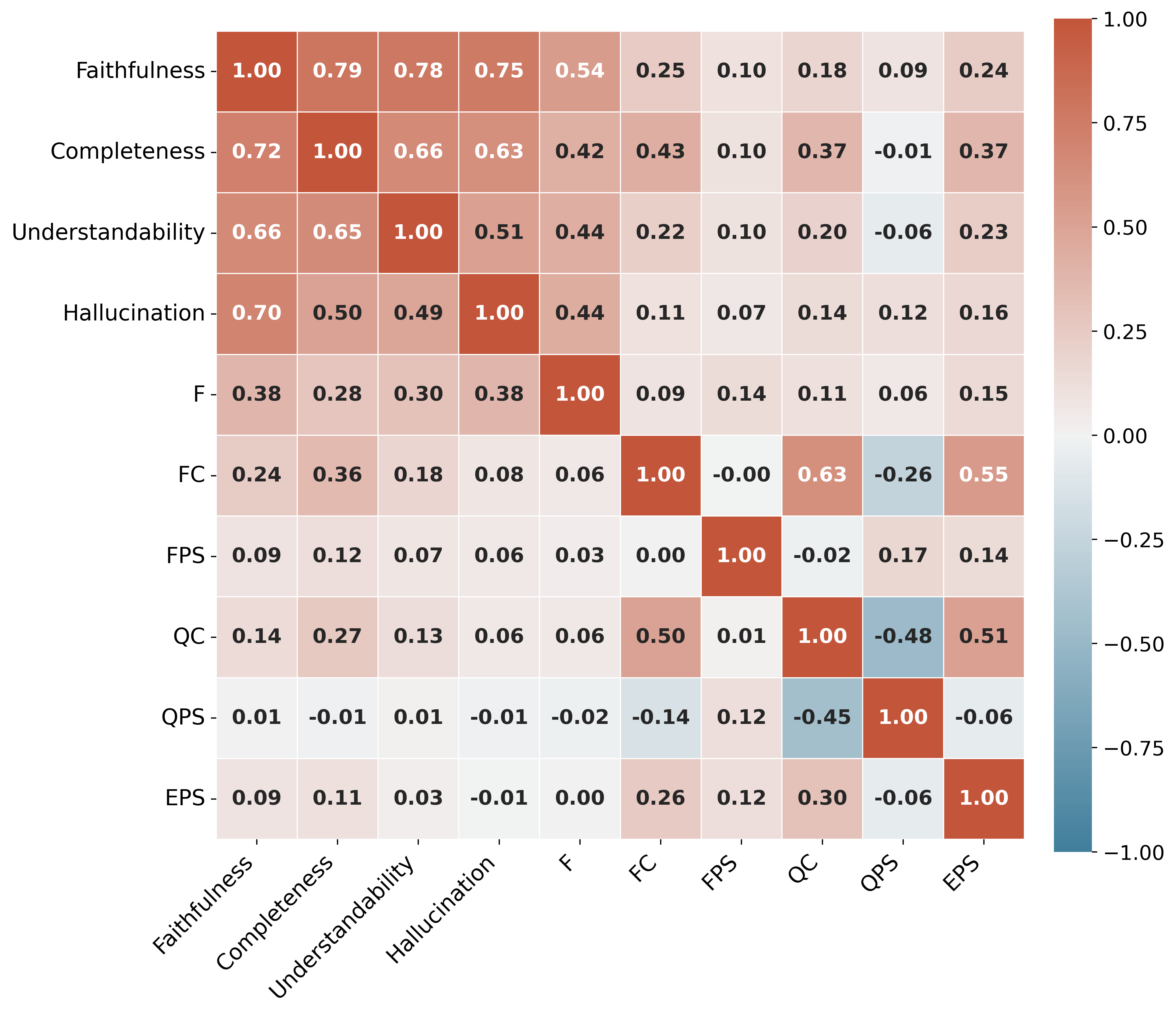}
\caption{Pairwise correlations of evaluation metrics on the Screenshots dataset. The upper triangle reports Spearman's $\rho$; the lower triangle reports Pearson's $r$.}
\label{fig:correlation}
\end{figure}

FPS, QPS, and EPS are only weakly correlated with one another (Figure~\ref{fig:correlation}), confirming they measure distinct aspects of summary quality.

The negative Pearson QPS/QC correlation ($r = -0.45$) reflects their differing objectives: QC rewards breadth of question coverage, while the $(1-R_s/R_c)$ term in QPS explicitly discounts it in favour of main questions.

\paragraph{A small model leads on two scaffold axes.}
On Contracts, GPT-4.1-mini outscores every larger model on both
QPS (0.75) and EPS (0.46) while sitting at parity on the holistic
axes. A team trusting holistic scores alone would pay for a bigger
model that is structurally worse, over-provisioning compute for a
task the smaller, cheaper model already wins. Saturation is thus
not only an evaluation problem; it translates directly into wasted
inference cost and weaker production decisions.

\paragraph{Humans aren't the ceiling; they're the target.}
Seven of eight models exceed the Human row on EPS (Human $= 0.76$). EPS is recall-oriented over scaffold attributes, an objective annotators are not writing toward, so the human row functions as a \emph{sensible-target} anchor rather than a ceiling, and is best read alongside the holistic axes rather than as a standalone quality verdict.

Scaffold metrics on public benchmarks, under grouped summarization, are reported in Table~\ref{tab:scaffold_grouped} and Figure~\ref{fig:scaffold_spread}.

\section{Failure-Mode Analysis}

Saturation is the most visible failure of holistic evaluation, but
not the only one. By constructing grouped multi-document settings from single-document benchmarks~\citep{Mukherjee2022ECTSumAN,Kornilova2019BillSumAC,Huang2021EfficientAF},
we observe four failure patterns in our data, F1 to F4
(Table~\ref{tab:failure_modes}). These are descriptive stress tests, not
isolated causal experiments. We summarize them here; details are in
Appendix~\ref{app:failure_modes}.

\begin{table*}[!t]
\centering\small
\setlength{\tabcolsep}{5pt}
\begin{tabularx}{\textwidth}{@{}L{0.45cm}L{2.45cm}Y L{3.85cm}@{}}
\toprule
ID & Name & Effect & Production consequence \\
\midrule
\rowcolor{warnamber}
F1 & Catastrophic compression collapse &
Hallucination drops from near-perfect (10.0) to 3.6 - 7.0 on the
same source when the judge conflates omission with fabrication &
\textbf{Good summaries look hallucinated} \\

\addlinespace
F2 & Ranking instability &
ROUGE-L and Direct Scoring select different best models in all four
tested (dataset, setting) configurations &
\textbf{Metric choice, not model quality, decides what ships} \\

\addlinespace
\rowcolor{warnamber}
F3 & Domain-driven ROUGE bias &
ROUGE-L scores are approximately $2\times$ higher on government
reports than earnings calls for summaries of comparable quality &
\textbf{Cross-domain comparison becomes invalid} \\

\addlinespace
F4 & Goodhart rule-breaking &
A model violating the 100-word budget achieves the highest grouped
direct score; prompt relaxation fails to correct the behavior &
\textbf{Non-compliant summaries get rewarded} \\

\bottomrule
\end{tabularx}

\caption{Four failure modes in grouped multi-document
evaluation, each paired with the
model-selection decision it corrupts.}

\label{tab:failure_modes}
\end{table*}

\begin{figure*}[!t]
\centering
\includegraphics[width=0.85\textwidth]
{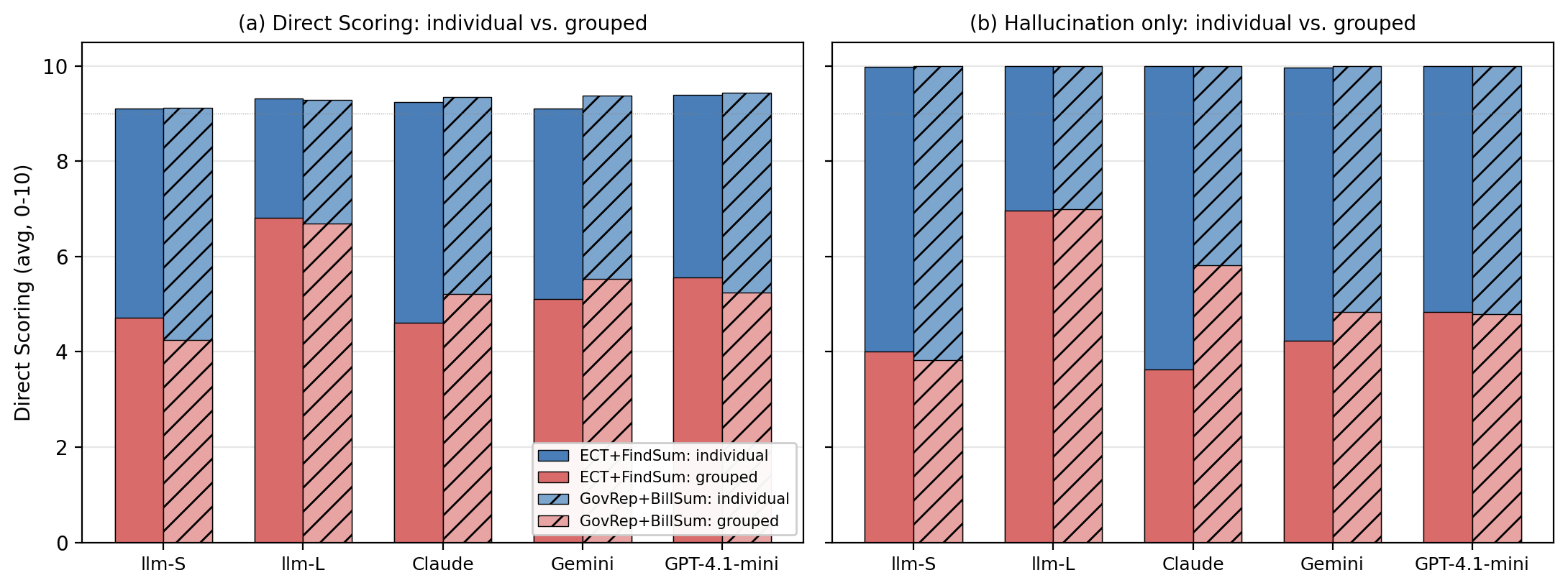}
\caption{\textbf{F1: compression collapse} on two multi-document benchmarks,
individual (\textit{blue}) versus grouped 100-word (\textit{red}). Direct
Scoring and Hallucination both decrease}
\label{fig:grouped_collapse}
\end{figure*}

\textbf{F1 (Catastrophic collapse under compression).} When five models compress
related documents into a 100-word budget, Direct Scoring falls from
\hlnum{9.1--9.4} individually to \hlsat{4.2--6.8} grouped, and Hallucination
decreases by about 5 points on average (Figure~\ref{fig:grouped_collapse}).
The cause is not fabrication but a measurement artifact: the judge compares the 100-word summary against a much longer reference, 
so any detail dropped to meet the budget looks ``unsupported.''

\begin{table*}[t]
\centering
\small
\setlength{\tabcolsep}{8pt}
\begin{tabularx}{\linewidth}{lXXX}
\toprule
\textbf{Dataset} &
\textbf{Set.} &
\textbf{ROUGE-L leader} &
\textbf{DS leader} \\
\midrule
ECT+FindSum & individual & \cellcolor{scaffoldblue!40}Claude-3.7-Sonnet (.111) & \cellcolor{warnamber}GPT-4.1-mini (9.40) \\
GovRep+BillSum & individual & \cellcolor{scaffoldblue!40}Gemini-2.5-Flash (.228) & \cellcolor{warnamber}GPT-4.1-mini (9.44) \\
ECT+FindSum & grouped & \cellcolor{scaffoldblue!40}GPT-4.1-mini (.085) & \cellcolor{warnamber}Internal-L (6.81) \\
GovRep+BillSum & grouped & \cellcolor{scaffoldblue!40}GPT-4.1-mini (.119) & \cellcolor{warnamber}Internal-L (6.69) \\
\bottomrule
\end{tabularx}
\caption{
F2 (\textbf{ranking instability}): ROUGE-L and Direct Scoring select
different metric leaders in all four (dataset, setting) configurations. No independent preference target establishes which leader has higher quality.
ECT\,=\,ECTSum~\citep{Mukherjee2022ECTSumAN},
FindSum~\citep{liu2022long},
GovRep\,=\,GovReport~\citep{Huang2021EfficientAF},
BillSum~\citep{Kornilova2019BillSumAC}.
}
\label{tab:rank_disagree_extended}
\end{table*}

The next three failures concern the rankings these metrics produce.
\textbf{F2 (instability):} ROUGE-L and Direct Scoring crown
different top models in all four (dataset, setting) configurations
we tried (Table~\ref{tab:rank_disagree_extended}); with disagreement that
complete, it is the choice of metric, not the quality of the model,
that decides which system ships. \textbf{F3 (domain bias):} for
summaries of comparable quality, ROUGE-L runs about \hlsat{$2\times$}
higher on government reports than on earnings calls, so a score is
only meaningful within a domain and cross-domain comparisons are
not valid. \textbf{F4 (Goodhart):} the model that most often breaks
the 100-word limit earns the highest grouped Direct Score, and
relaxing the limit only lowers ROUGE-L without fixing the behavior.

\paragraph{Why hierarchical scaffolds help.}

Hierarchical scaffold addresses this by scoring components independently. On the grouped dataset, scaffold metrics sustain separation (Supporting Fact Coverage: \hlbest{$3.4\times$}, 
Main Fact Coverage: \hlbest{$1.8\times$}; Appendix~\ref{app:scaffold_grouped}, Table~\ref{tab:scaffold_grouped}), exposing the real culprit: the Hallucination collapse isn't fabrication, models drop supporting details under 
compression. While holistic scores flag that something's wrong, scaffolds reveal the cause.

\section{Conclusion}

When every frontier model earns a near-perfect hallucination score, the bottleneck has shifted from the models to the way we measure them. We argued that holistic LLM-judge metrics lose discriminative power exactly as generation quality rises, and we showed it across two proprietary datasets, three public benchmarks, and ten languages, then cataloged four recurring failure modes of the conventional stack.

Our response, the \textbf{Semantic Scaffold}, changes the structure of evaluation rather than the prompt: it extracts a hierarchical reference from the source once and scores every candidate against it, so the scores trace back to specific facts, questions, and entities. Where holistic axes flatten, its FPS, QPS, and EPS metrics still resolve differences between models, and under heavy compression they pinpoint \emph{why} a model fails instead of collapsing into uninterpretable numbers. \\
\indent \textbf{Takeaway}: when every model scores near the ceiling, the task is probably not solved; the metric has stopped resolving, and that is the moment to switch to structural, decomposed evaluation.

\bibliography{colm2026_conference}

@inproceedings{lin2004rouge,
  title={Rouge: A package for automatic evaluation of summaries},
  author={Lin, Chin-Yew},
  booktitle={Text summarization branches out},
  pages={74--81},
  year={2004},
  url={https://aclanthology.org/W04-1013/}
}

@inproceedings{papineni2002bleu,
  title={Bleu: a method for automatic evaluation of machine translation},
  author={Papineni, Kishore and Roukos, Salim and Ward, Todd and Zhu, Wei-Jing},
  booktitle={Proceedings of the 40th annual meeting of the Association for Computational Linguistics},
  pages={311--318},
  year={2002},
  doi={10.3115/1073083.1073135},
  url={https://aclanthology.org/P02-1040/}
}

@inproceedings{banerjee2005meteor,
  title={METEOR: An automatic metric for MT evaluation with improved correlation with human judgments},
  author={Banerjee, Satanjeev and Lavie, Alon},
  booktitle={Proceedings of the acl workshop on intrinsic and extrinsic evaluation measures for machine translation and/or summarization},
  pages={65--72},
  year={2005},
  url={https://aclanthology.org/W05-0909/}
}

@inproceedings{zhang2019bertscore,
  title={{BERTScore}: Evaluating Text Generation with {BERT}},
  author={Zhang, Tianyi and Kishore, Varsha and Wu, Felix and Weinberger, Kilian Q and Artzi, Yoav},
  booktitle={International Conference on Learning Representations ({ICLR})},
  year={2020},
  url={https://openreview.net/forum?id=SkeHuCVFDr}
}

@inproceedings{zhao2019moverscore,
  title={MoverScore: Text generation evaluating with contextualized embeddings and earth mover distance},
  author={Zhao, Wei and Peyrard, Maxime and Liu, Fei and Gao, Yang and Meyer, Christian M and Eger, Steffen},
  booktitle={Proceedings of the 2019 conference on empirical methods in natural language processing and the 9th international joint conference on natural language processing (EMNLP-IJCNLP)},
  pages={563--578},
  year={2019},
  doi={10.18653/v1/D19-1053},
  url={https://aclanthology.org/D19-1053/}
}

@inproceedings{kryscinski2020evaluating,
  title={Evaluating the factual consistency of abstractive text summarization},
  author={Kry{\'s}ci{\'n}ski, Wojciech and McCann, Bryan and Xiong, Caiming and Socher, Richard},
  booktitle={Proceedings of the 2020 conference on empirical methods in natural language processing (EMNLP)},
  pages={9332--9346},
  year={2020},
  doi={10.18653/v1/2020.emnlp-main.750},
  url={https://aclanthology.org/2020.emnlp-main.750/}
}

@article{laban2022summac,
  title={{SummaC}: Re-visiting {NLI}-based Models for Inconsistency Detection in Summarization},
  author={Laban, Philippe and Schnabel, Tobias and Bennett, Paul N. and Hearst, Marti A.},
  journal={Transactions of the Association for Computational Linguistics},
  volume={10},
  pages={163--177},
  year={2022},
  doi={10.1162/tacl_a_00453},
  url={https://aclanthology.org/2022.tacl-1.10/}
}

@article{deutsch2021towards,
  title={Towards Question-Answering as an Automatic Metric for Evaluating the Content Quality of a Summary},
  author={Deutsch, Daniel and Bedrax-Weiss, Tania and Roth, Dan},
  journal={Transactions of the Association for Computational Linguistics},
  volume={9},
  pages={774--789},
  year={2021},
  doi={10.1162/tacl_a_00397},
  url={https://aclanthology.org/2021.tacl-1.47/}
}

@inproceedings{scialom2021questeval,
  title={QuestEval: Summarization asks for fact-based evaluation},
  author={Scialom, Thomas and Dray, Paul-Alexis and Lamprier, Sylvain and Piwowarski, Benjamin and Staiano, Jacopo and Wang, Alex and Gallinari, Patrick},
  booktitle={Proceedings of the 2021 conference on empirical methods in natural language processing},
  pages={6594--6604},
  year={2021},
  doi={10.18653/v1/2021.emnlp-main.529},
  url={https://aclanthology.org/2021.emnlp-main.529/}
}

@inproceedings{liu2023g,
  title={G-eval: NLG evaluation using gpt-4 with better human alignment},
  author={Liu, Yang and Iter, Dan and Xu, Yichong and Wang, Shuohang and Xu, Ruochen and Zhu, Chenguang},
  booktitle={Proceedings of the 2023 conference on empirical methods in natural language processing},
  pages={2511--2522},
  year={2023},
  doi={10.18653/v1/2023.emnlp-main.153},
  url={https://aclanthology.org/2023.emnlp-main.153/}
}

@inproceedings{gabriel2021go,
  title={{GO FIGURE}: A Meta Evaluation of Factuality in Summarization},
  author={Gabriel, Saadia and Celikyilmaz, Asli and Jha, Rahul and Choi, Yejin and Gao, Jianfeng},
  booktitle={Findings of the Association for Computational Linguistics: ACL-IJCNLP 2021},
  pages={478--487},
  year={2021},
  doi={10.18653/v1/2021.findings-acl.42},
  url={https://aclanthology.org/2021.findings-acl.42/}
}

@article{fabbri2021summeval,
  title={{SummEval}: Re-evaluating Summarization Evaluation},
  author={Fabbri, Alexander R. and Kry{\'s}ci{\'n}ski, Wojciech and McCann, Bryan and Xiong, Caiming and Socher, Richard and Radev, Dragomir},
  journal={Transactions of the Association for Computational Linguistics},
  volume={9},
  pages={391--409},
  year={2021},
  doi={10.1162/tacl_a_00373},
  url={https://aclanthology.org/2021.tacl-1.24/}
}

@article{Goyal2022NewsSA,
  title={News Summarization and Evaluation in the Era of {GPT-3}},
  author={Goyal, Tanya and Li, Junyi Jessy and Durrett, Greg},
  journal={arXiv preprint arXiv:2209.12356},
  year={2022},
  doi={10.48550/arXiv.2209.12356},
  url={https://arxiv.org/abs/2209.12356}
}

@article{Zhang2023BenchmarkingLL,
  title={Benchmarking Large Language Models for News Summarization},
  author={Zhang, Tianyi and Ladhak, Faisal and Durmus, Esin and Liang, Percy and McKeown, Kathleen and Hashimoto, Tatsunori B.},
  journal={Transactions of the Association for Computational Linguistics},
  year={2024},
  volume={12},
  pages={39--57},
  doi={10.1162/tacl_a_00632},
  url={https://aclanthology.org/2024.tacl-1.3/}
}

@inproceedings{See2017GetTT,
  title={Get To The Point: Summarization with Pointer-Generator Networks},
  author={See, Abigail and Liu, Peter J. and Manning, Christopher D.},
  booktitle={Proceedings of the 55th Annual Meeting of the Association for Computational Linguistics ({ACL})},
  pages={1073--1083},
  year={2017},
  doi={10.18653/v1/P17-1099},
  url={https://aclanthology.org/P17-1099/}
}

@inproceedings{song2024finesure,
  title={{FineSurE}: Fine-grained Summarization Evaluation using {LLMs}},
  author={Song, Hwanjun and Su, Hang and Shalyminov, Igor and Cai, Jason and Mansour, Saab},
  booktitle={Proceedings of the 62nd Annual Meeting of the Association for Computational Linguistics (Volume 1: Long Papers)},
  pages={906--922},
  year={2024},
  doi={10.18653/v1/2024.acl-long.51},
  url={https://aclanthology.org/2024.acl-long.51/}
}

@inproceedings{nenkova-passonneau-2004-evaluating,
    title = "Evaluating Content Selection in Summarization: The Pyramid Method",
    author = "Nenkova, Ani  and
      Passonneau, Rebecca",
    booktitle = "Proceedings of the Human Language Technology Conference of the North {A}merican Chapter of the Association for Computational Linguistics: {HLT}-{NAACL} 2004",
    month = may # " 2 - " # may # " 7",
    year = "2004",
    address = "Boston, Massachusetts, USA",
    publisher = "Association for Computational Linguistics",
    url = "https://aclanthology.org/N04-1019/",
    pages = "145--152"
}

@inproceedings{Min2023FActScoreFA,
  title={{FActScore}: Fine-grained Atomic Evaluation of Factual Precision in Long Form Text Generation},
  author={Min, Sewon and Krishna, Kalpesh and Lyu, Xinxi and Lewis, Mike and Yih, Wen-tau and Koh, Pang Wei and Iyyer, Mohit and Zettlemoyer, Luke and Hajishirzi, Hannaneh},
  booktitle={Proceedings of the 2023 Conference on Empirical Methods in Natural Language Processing ({EMNLP})},
  pages={12076--12100},
  year={2023},
  doi={10.18653/v1/2023.emnlp-main.741},
  url={https://aclanthology.org/2023.emnlp-main.741/}
}

@inproceedings{Liu2022RevisitingTG,
  title={Revisiting the Gold Standard: Grounding Summarization Evaluation with Robust Human Evaluation},
  author={Liu, Yixin and Fabbri, Alexander R. and Liu, Pengfei and Zhao, Yilun and Nan, Linyong and Han, Ruilin and Han, Simeng and Joty, Shafiq R. and Wu, Chien-Sheng and Xiong, Caiming and Radev, Dragomir R.},
  booktitle={Proceedings of the 61st Annual Meeting of the Association for Computational Linguistics ({ACL})},
  pages={4140--4170},
  year={2023},
  doi={10.18653/v1/2023.acl-long.228},
  url={https://aclanthology.org/2023.acl-long.228/}
}

@inproceedings{Liu2023TowardsIA,
  title={Towards Interpretable and Efficient Automatic Reference-Based Summarization Evaluation},
  author={Liu, Yixin and Fabbri, Alexander R. and Zhao, Yilun and Liu, Pengfei and Joty, Shafiq R. and Wu, Chien-Sheng and Xiong, Caiming and Radev, Dragomir R.},
  booktitle={Proceedings of the 2023 Conference on Empirical Methods in Natural Language Processing ({EMNLP})},
  pages={16360--16368},
  year={2023},
  doi={10.18653/v1/2023.emnlp-main.1018},
  url={https://aclanthology.org/2023.emnlp-main.1018/}
}

@inproceedings{Tang2024MiniCheckEF,
  title={{MiniCheck}: Efficient Fact-Checking of {LLMs} on Grounding Documents},
  author={Tang, Liyan and Laban, Philippe and Durrett, Greg},
  booktitle={Proceedings of the 2024 Conference on Empirical Methods in Natural Language Processing ({EMNLP})},
  pages={8818--8847},
  year={2024},
  doi={10.18653/v1/2024.emnlp-main.499},
  url={https://aclanthology.org/2024.emnlp-main.499/}
}

@inproceedings{lee2025towards,
  title={Towards a Holistic and Automated Evaluation Framework for Multi-Level Comprehension of LLMs in Book-Length Contexts},
  author={Lee, Yuho and Deng, Jiaqi and Kim, Nicole Hee-Yeon and Min, Hyangsuk and Yun, Taewon and Ban, Minjeong and Yul, Kim and Song, Hwanjun},
  booktitle={Proceedings of the 2025 Conference on Empirical Methods in Natural Language Processing},
  pages={24401--24425},
  year={2025},
  doi={10.18653/v1/2025.emnlp-main.1241},
  url={https://aclanthology.org/2025.emnlp-main.1241/}
}

@inproceedings{wan2024acueval,
  title={{ACUEval}: Fine-grained Hallucination Evaluation and Correction for Abstractive Summarization},
  author={Wan, David and Sinha, Koustuv and Iyer, Srini and Celikyilmaz, Asli and Bansal, Mohit and Pasunuru, Ramakanth},
  booktitle={Findings of the Association for Computational Linguistics: ACL 2024},
  pages={10036--10056},
  year={2024},
  doi={10.18653/v1/2024.findings-acl.597},
  url={https://aclanthology.org/2024.findings-acl.597/}
}

@inproceedings{Zha2023AlignScoreEF,
  title={{AlignScore}: Evaluating Factual Consistency with a Unified Alignment Function},
  author={Zha, Yuheng and Yang, Yichi and Li, Ruichen and Hu, Zhiting},
  booktitle={Proceedings of the 61st Annual Meeting of the Association for Computational Linguistics ({ACL})},
  pages={11328--11348},
  year={2023},
  doi={10.18653/v1/2023.acl-long.634},
  url={https://aclanthology.org/2023.acl-long.634/}
}

@inproceedings{Sharma2019BIGPATENTAL,
  title={{BIGPATENT}: A Large-Scale Dataset for Abstractive and Coherent Summarization},
  author={Sharma, Eva and Li, Chen and Wang, Lu},
  booktitle={Proceedings of the 57th Annual Meeting of the Association for Computational Linguistics ({ACL})},
  pages={2204--2213},
  year={2019},
  doi={10.18653/v1/P19-1212},
  url={https://aclanthology.org/P19-1212/}
}

@inproceedings{Cohan2018ADA,
  title={A Discourse-Aware Attention Model for Abstractive Summarization of Long Documents},
  author={Cohan, Arman and Dernoncourt, Franck and Kim, Doo Soon and Bui, Trung and Kim, Seokhwan and Chang, Walter and Goharian, Nazli},
  booktitle={Proceedings of the 2018 Conference of the North American Chapter of the Association for Computational Linguistics ({NAACL-HLT})},
  pages={615--621},
  year={2018},
  doi={10.18653/v1/N18-2097},
  url={https://aclanthology.org/N18-2097/}
}

@inproceedings{Huang2021EfficientAF,
  title={Efficient Attentions for Long Document Summarization},
  author={Huang, Luyang and Cao, Shuyang and Parulian, Nikolaus and Ji, Heng and Wang, Lu},
  booktitle={Proceedings of the 2021 Conference of the North American Chapter of the Association for Computational Linguistics ({NAACL-HLT})},
  pages={1419--1436},
  year={2021},
  doi={10.18653/v1/2021.naacl-main.112},
  url={https://aclanthology.org/2021.naacl-main.112/}
}

@inproceedings{Mukherjee2022ECTSumAN,
  title={{ECTSum}: A New Benchmark Dataset For Bullet Point Summarization of Long Earnings Call Transcripts},
  author={Mukherjee, Rajdeep and Bohra, Abhinav and Banerjee, Akash and Sharma, Soumya and Hegde, Manjunath and Shaikh, Afreen and Shrivastava, Shivani and Dasgupta, Koustuv and Ganguly, Niloy and Ghosh, Saptarshi and Goyal, Pawan},
  booktitle={Proceedings of the 2022 Conference on Empirical Methods in Natural Language Processing ({EMNLP})},
  pages={10893--10906},
  year={2022},
  doi={10.18653/v1/2022.emnlp-main.748},
  url={https://aclanthology.org/2022.emnlp-main.748/}
}

@inproceedings{Kornilova2019BillSumAC,
  title={{BillSum}: A Corpus for Automatic Summarization of {US} Legislation},
  author={Kornilova, Anastassia and Eidelman, Vladimir},
  booktitle={Proceedings of the 2nd Workshop on New Frontiers in Summarization},
  pages={48--56},
  year={2019},
  doi={10.18653/v1/D19-5406},
  url={https://aclanthology.org/D19-5406/}
}

@inproceedings{Zheng2023JudgingLW,
  title={Judging {LLM}-as-a-Judge with {MT-Bench} and Chatbot Arena},
  author={Zheng, Lianmin and Chiang, Wei-Lin and Sheng, Ying and Zhuang, Siyuan and Wu, Zhanghao and Zhuang, Yonghao and Lin, Zi and Li, Zhuohan and Li, Dacheng and Xing, Eric P. and Zhang, Hao and Gonzalez, Joseph E. and Stoica, Ion},
  booktitle={Advances in Neural Information Processing Systems ({NeurIPS})},
  volume={36},
  year={2023},
  url={https://proceedings.neurips.cc/paper_files/paper/2023/hash/91f18a1287b398d378ef22505bf41f6d-Abstract-Datasets_and_Benchmarks.html}
}

@inproceedings{liu2022long,
  title={Long Text and Multi-Table Summarization: Dataset and Method},
  author={Liu, Shuaiqi and Cao, Jiannong and Yang, Ruosong and Wen, Zhiyuan},
  booktitle={Findings of the Association for Computational Linguistics: EMNLP 2022},
  pages={1995--2010},
  year={2022},
  doi={10.18653/v1/2022.findings-emnlp.145},
  url={https://aclanthology.org/2022.findings-emnlp.145/}
}
\bibliographystyle{colm2026_conference}

\section*{Limitations and Future Work}
\label{ssec:limitations}

The main limitation of our framework is its reliance on the quality of the initial semantic scaffold. 
All downstream metrics depend on the accuracy and completeness of the LLM-based extraction. 
Therefore, a key direction for future work is to develop stronger methods to 
improve and automatically assess scaffold quality, especially in specialized or noisy domains. 
Further research will also explore broader uses of the Semantic Scaffold, 
such as content recommendation, educational assessment, and cross-lingual analysis.

Two extensions would strengthen the framework's use as a selection criterion. First, a preference-based meta-evaluation: Screenshots retains three human references per item, making it well suited to relating scaffold scores to human preference and to calibrating the main/supporting labels against human salience judgments. Second, broader reporting: pairwise comparison is the standard non-saturating alternative to direct scoring, and situating the scaffold metrics against it, with length normalization and interval estimates, would clarify where each is preferable.

\section*{Ethics Statement}

The proprietary \textbf{Screenshots} and \textbf{Contracts}
datasets were annotated in-house under standard employment
agreements; no crowd-sourcing was used. Source documents and
scaffolds exclude end-user personally identifiable information
(PII), and entity annotations cover organizations, products,
system identifiers, and document-internal references rather than
personal attributes.

Because scaffold construction and scoring are LLM-assisted, the
resulting metrics should be interpreted as comparative evaluation
signals rather than absolute guarantees of summary quality, especially
in sensitive use cases.

\appendix

\section{Semantic Scaffold Evaluation Framework: Metric Overview}
\label{app:metrics_summary}

Table~\ref{tab:metrics_summary} summarizes the key metrics used in our
framework, organized into Baseline, Foundational, and Showcase
categories. The Foundational metrics (Section~\ref{sec:method},
Table~\ref{tab:foundational}) assess basic factual accuracy and
coverage, while the Showcase metrics (FPS, QPS, EPS) evaluate the
higher-order structural and entity-level qualities that remain
informative once holistic LLM-judge scores saturate. The subsequent
sections give the exact prompt templates for each.

\begin{figure*}[!h]
\centering
\begin{minipage}{0.45\textwidth}
  \centering
  \includegraphics[width=\textwidth]{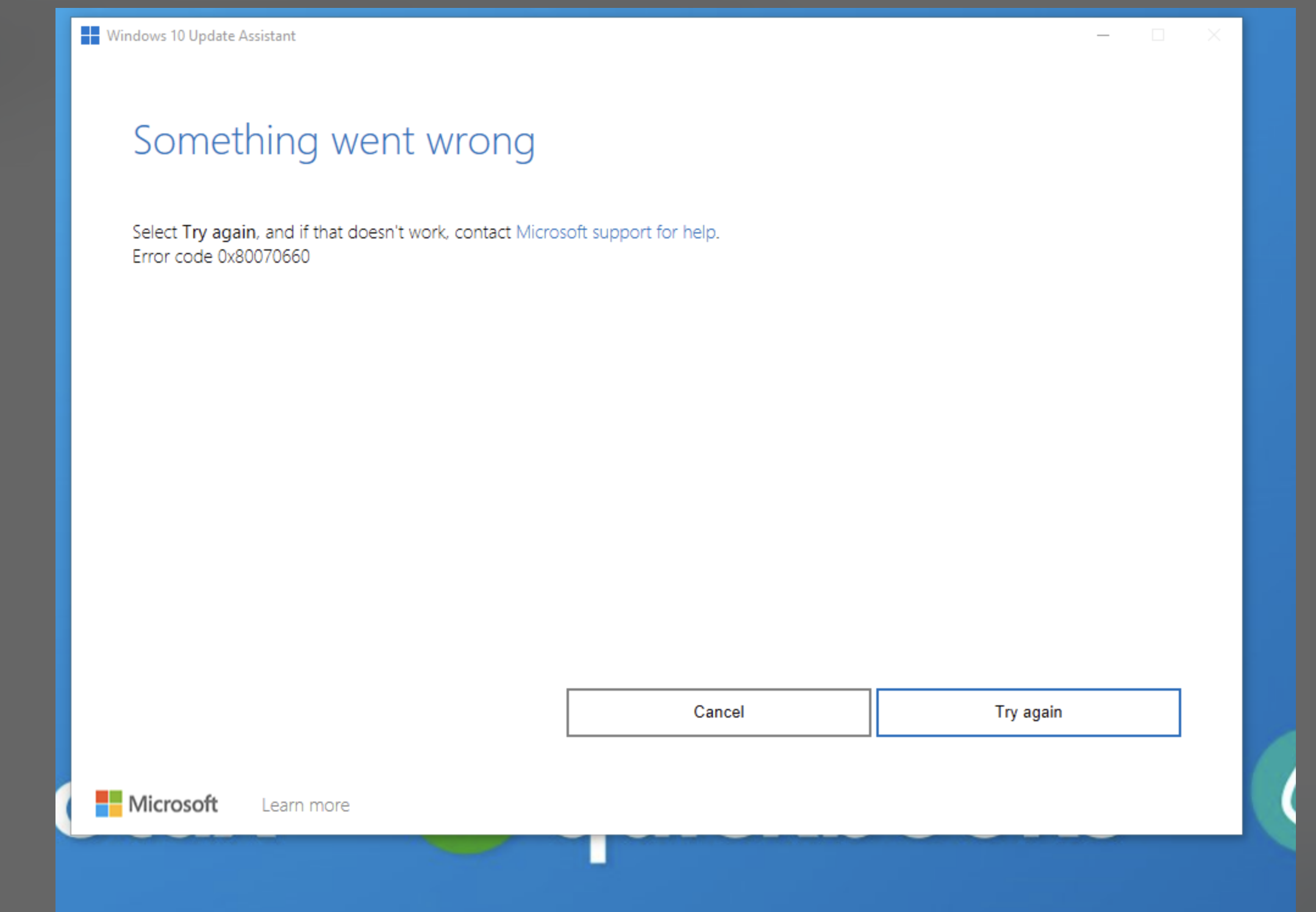}
\end{minipage}\hfill
\begin{minipage}{0.45\textwidth}
  \centering
  \includegraphics[width=\textwidth]{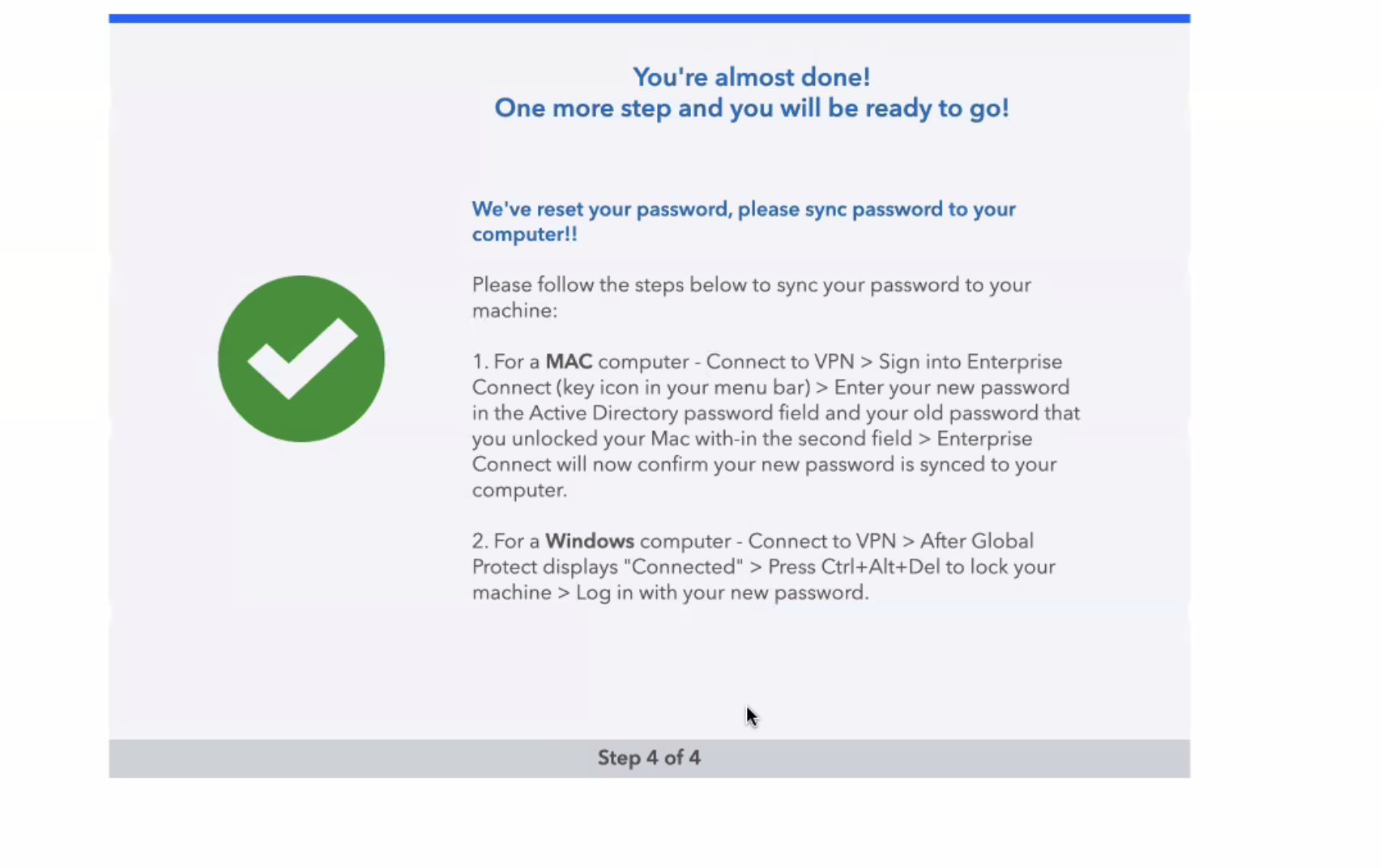}
\end{minipage}
\caption{\textbf{Screenshot} examples from the proprietary dataset:
a Windows Update error dialog (\textit{left}) and a password-sync
instruction page (\textit{right}). Error codes, UI labels, and
step-by-step identifiers drive summary quality, making \textbf{EPS}
differences salient.}
\label{fig:screenshots}
\end{figure*}

\begin{table*}[t]
\centering
\small
\setlength{\tabcolsep}{5pt}
\renewcommand{\arraystretch}{1.4}
\begin{tabularx}{\textwidth}{@{}L{3.35cm}L{3.0cm}Y@{}}
\toprule
\textbf{Metric} & \textbf{Measures} & \textbf{Description} \\
\midrule
\rowcolor{saturatepink!20}\multicolumn{3}{l}{\textbf{\textit{Baseline}}} \\
LLM-as-Judge & Holistic Quality & LLM rates summary on faithfulness, completeness, coherence, relevance. Susceptible to biases. \\
\midrule
\multicolumn{3}{l}{\textbf{\textit{Foundational: Fact-Based}}} \\
Fact Faithfulness (F) & Fact Accuracy & Proportion of key facts accurately represented. \\
Fact Coverage (FC) & Completeness & Proportion of key facts present in summary. \\
Main Fact Coverage (Main FC) & Main-point Recall & $M_s/M_c$, the proportion of main facts present in the summary. \\
Supporting Fact Coverage (Supp.\ FC) & Supporting-detail Recall & $S_s/S_c$, the proportion of supporting facts present in the summary. \\
Fact Focus (F.\ Focus) & Factual Precision & Proportion of summary sentences aligned to at least one scaffold fact. \\
\midrule
\multicolumn{3}{l}{\textbf{\textit{Foundational: Question-Based}}} \\
Question Coverage (QC) & Completeness & Proportion of key questions answered in summary. \\
\midrule
\rowcolor{scaffoldblue!30}\multicolumn{3}{l}{\textbf{\textit{Showcase Metrics: Hierarchy}}} \\
Fact Preservation (FPS) & Information Hierarchy & Preserves main facts while avoiding overemphasis on supporting details. \\
Question Preservation (QPS) & Hierarchical QA & Answers main questions without excessive focus on supporting ones. \\
Entity Preservation (EPS) & Attribute Accuracy & Correctly preserves entity-attribute pairs from source. \\
\bottomrule
\end{tabularx}
\caption{Evaluation metrics in the Semantic Scaffold framework.}
\label{tab:metrics_summary}
\end{table*}

\section{Qualitative Walkthrough: From Source to Scaffold to Scores}
\label{app:walkthrough}

This appendix walks one example
(\texttt{internal-dataset-v1/task-0099}) from raw OCR text through
scaffold construction to final scores, as a concrete reference for
reproducing the pipeline.

\subsection{Source Document (OCR)}

\begin{quote}
\small\itshape
Camera. Allow access to the camera on this device. If you allow
access, people using this device will be able to choose if their
apps have camera access by using the settings on this page.
Denying access blocks Microsoft Store apps and most desktop
apps from accessing the camera. It does not block Windows Hello.
Camera access for this device is on. Change. Allow apps to
access your camera. If you allow access, you can choose which
apps can access your camera by using the settings on this page.
Denying access blocks apps from accessing your camera. It does
not block Windows Hello. On. Some desktop apps may still be able
to access your camera when settings on this page are off.
Find out why.
\end{quote}

These 108 OCR words from a Windows~10 privacy-settings screenshot pack
a two-level permission hierarchy, two exceptions (\emph{Windows Hello}
and desktop apps), a state (access enabled), and an action
(``Find out why''). It is a typical \textbf{Screenshots} sample:
short and entity-rich, where small differences in factual and entity
preservation separate strong summaries from adequate ones and where
holistic metrics blur exactly what the scaffold keeps.

\subsection{Extracted Scaffold}

The scaffold organizes the document into three views (atomic
facts, key questions, and entity attributes), each item labeled
\emph{main point} or \emph{supporting detail}.

\noindent\textbf{Atomic Facts}
(main points first, followed by supporting details).

\begin{enumerate}\small
\item[M1.]
Camera access on the device can be enabled or disabled.
\textit{(main point)}

\item[M2.]
When device-level access is enabled, users can choose which
individual applications are permitted camera access.
\textit{(main point)}

\item[M3.]
Denying device-level access blocks Microsoft Store applications
and most desktop applications from accessing the camera.
\textit{(main point)}

\item[M4.]
Camera access for this device is currently enabled.
\textit{(main point)}

\item[S1.]
Denying camera access does not block
\emph{Windows Hello}.
\textit{(supporting detail)}

\item[S2.]
Some desktop applications may still access the camera even when
per-application settings are disabled.
\textit{(supporting detail)}

\item[S3.]
A ``Find out why'' link is provided to explain the desktop
application exception.
\textit{(supporting detail)}
\end{enumerate}

\noindent\textbf{Key Questions}
(main questions first, followed by supporting details).

\begin{enumerate}\small
\item[Q1.]
What functionality does enabling device-level camera access
provide to users?
\textit{(main point)}

\item[Q2.]
What happens to application-level camera access when
device-level access is denied?
\textit{(main point)}

\item[Q3.]
What is the current camera access state for this device?
\textit{(main point)}

\item[R1.]
Does denying camera access affect
\emph{Windows Hello}?
\textit{(supporting detail)}

\item[R2.]
Can desktop applications still access the camera when
per-application settings are disabled?
\textit{(supporting detail)}
\end{enumerate}

\noindent\textbf{Entity Attributes}

\begin{itemize}\small

\item \textbf{Camera access (device level)}
\begin{itemize}\small
    \item Can be enabled or disabled
    \item Currently enabled
    \item Enables per-application access control when active
\end{itemize}

\item \textbf{Camera access (application level)}
\begin{itemize}\small
    \item Individual applications may be permitted or blocked
    \item Per-application settings are available only when
    device-level access is enabled
\end{itemize}

\item \textbf{Windows Hello}
\begin{itemize}\small
    \item Not blocked by camera access denial
    \item Operates independently of the privacy settings page
\end{itemize}

\item \textbf{Desktop applications}
\begin{itemize}\small
    \item Blocked when device-level access is denied
    \item May still access the camera despite
    per-application settings being disabled
\end{itemize}

\end{itemize}

\subsection{Model Summaries and Per-Instance Scores}

We compare four models on the same instance. All four are
broadly faithful and coherent, yet they differ in how well they
preserve the document's structure, entity attributes, and
exceptions.

\noindent\textbf{\texttt{Internal-S}}
(189 characters):
\begin{quote}
\small\itshape
``The document explains camera access settings for a device,
allowing users to choose which apps can access the camera, with
a note that denying access blocks most apps but not
Windows Hello.''
\end{quote}

\noindent\textbf{\texttt{Internal-M}}
(173 characters):
\begin{quote}
\small\itshape
``Camera access settings for this device and apps are detailed,
with options to allow or deny access.
Some desktop apps may still access the camera even when settings
are off.''
\end{quote}

\noindent\textbf{\texttt{Internal-L}}
(162 characters):
\begin{quote}
\small\itshape
``The document explains camera access settings, allowing or
denying access to the camera for apps, with notes on exceptions
for Windows Hello and some desktop apps.''
\end{quote}

\noindent\textbf{GPT-4o}
(223 characters):
\begin{quote}
\small\itshape
``The document explains camera access settings on a device,
allowing users to control app access to the camera.
It notes that denying access blocks most apps except
Windows Hello and mentions exceptions for some desktop apps.''
\end{quote}

\begin{table}[t]
\centering\small
\setlength{\tabcolsep}{5pt}
\begin{tabular}{lcccc}
\toprule
Model & \cellcolor{saturatepink!30}Faith. & \cellcolor{saturatepink!30}Hall. & \cellcolor{scaffoldblue!40}FPS & \cellcolor{scaffoldblue!40}EPS \\
\midrule
\texttt{Internal-S} & \cellcolor{saturatepink!15}0.92 & \cellcolor{saturatepink!15}0.98 & 0.33 & 0.44 \\
\texttt{Internal-M} & \cellcolor{saturatepink!15}0.90 & \cellcolor{saturatepink!15}0.97 & 0.22 & 0.33 \\
\texttt{Internal-L} & \cellcolor{saturatepink!15}0.93 & \cellcolor{saturatepink!15}0.98 & \cellcolor{bestgreen}0.38 & 0.56 \\
GPT-4o           & \cellcolor{saturatepink!15}0.94 & \cellcolor{saturatepink!15}0.99 & 0.25 & \cellcolor{bestgreen}0.67 \\
\bottomrule
\end{tabular}

\caption{
Illustrative per-instance scores for
\texttt{task-0099}
(approximate; final values depend on
sentence-level alignments produced by the
judge). Holistic metrics (Faithfulness,
Hallucination) stay clustered across all
four summaries, while FPS and EPS separate
them.}
\label{tab:qualitative_scores}
\end{table}

\subsection{Step-by-Step FPS and EPS Computation}

We walk through \texttt{Internal-S} to show that scaffold
scores trace to specific semantic units rather than an opaque judgment.

\paragraph{FPS.}
The scaffold has $M_c=4$ main facts (M1--M4) and $S_c=3$ supporting
facts (S1--S3). The summary covers M2 (``choose which apps can access
the camera'') and M3 (``denying access blocks most apps''), partially
signals M1, and omits M4, so $M_s=2$; it mentions S1 (``not Windows
Hello'') but not S2 or S3, so $S_s=1$. By Equation~\ref{eq:fps},
$\mathrm{FPS}=\frac{2}{4}\!\left(1-\frac{1}{3}\right)\approx0.33$.
GPT-4o covers more main facts ($M_s=3$) but reproduces more supporting
detail ($S_s=2$), giving
$\mathrm{FPS}=\frac{3}{4}\!\left(1-\frac{2}{3}\right)=0.25$: despite
higher raw coverage it scores lower, because the metric rewards
hierarchy, not volume.

\paragraph{EPS.}
The scaffold holds 9 entity attributes across 4 entities. The small
model preserves 2/3 device-level attributes, 1/2 application-level, 1/2
for Windows Hello, and 0/2 for desktop applications, so
$\mathrm{EPS}=4/9\approx0.44$, matching the automated judge.

\paragraph{Key takeaway.}
All four summaries get nearly identical holistic scores
(\hlsat{Faithfulness $\ge 0.90$}, \hlsat{Hallucination $\ge 0.97$}), yet FPS
spans about \hlnum{0.16} and EPS about \hlnum{0.34} on this single instance.
Such per-instance gaps compound across the dataset into the
larger separations in Table~\ref{tab:results}.

\section{Full Cross-Lingual Saturation Results}
\label{app:crosslingual}

\S\ref{sec:saturation} reports cross-lingual results averaged across datasets.
This appendix gives the full per-language breakdown for all three
evaluated models and datasets.

\paragraph{Evaluation setup.}
We test multilingual robustness on three internal screenshot
datasets: \textbf{Screen-v1} (887 images), \textbf{Dart-v1}
(831), and \textbf{Dart-qe-v1} (215). OCR text was professionally
translated into nine languages with human review. Summaries used
the same production prompt as English; to avoid English-pivot
bias, the judge prompt was translated too, so each summary is
scored against its target-language source. This measures
end-to-end multilingual quality, not translation fidelity alone.

\begin{table*}[t]
\centering
\scriptsize
\setlength{\tabcolsep}{2.5pt}

\resizebox{0.88\textwidth}{!}{
\begin{tabular}{lllcccccccccc}
\toprule
\textbf{Model} &
\textbf{Dataset} &
\textbf{Metric} &
\textbf{en} &
\textbf{de} &
\textbf{es} &
\textbf{fr} &
\textbf{it} &
\textbf{ja} &
\textbf{ko} &
\textbf{nl} &
\textbf{pt} &
\textbf{pt-BR} \\
\midrule

\multirow{6}{*}{Gemini-2.5-Flash}
& Dart-v1
& F
& 9.69 & 9.63 & 9.62 & 9.64 & 9.66
& 9.67 & 9.63 & 9.67 & 9.61 & 9.61 \\
&
& H
& 9.98 & 9.99 & 9.98 & 9.99 & 10.00
& 9.99 & 9.99 & 10.00 & 9.99 & 9.99 \\

& Dart-qe-v1
& F
& 9.67 & 9.61 & 9.58 & 9.60 & 9.61
& 9.65 & 9.57 & 9.64 & 9.59 & 9.62 \\
&
& H
& 9.99 & 9.98 & 9.99 & 9.98 & 10.00
& 9.99 & 9.98 & 9.98 & 9.99 & 9.99 \\

& Screen-v1
& F
& 9.65 & 9.64 & 9.62 & 9.63 & 9.60
& 9.64 & 9.58 & 9.65 & 9.62 & 9.62 \\
&
& H
& 9.99 & 9.99 & 9.99 & 9.99 & 9.99
& 9.98 & 9.99 & 9.99 & 9.99 & 9.99 \\
\midrule

\multirow{6}{*}{GPT-4.1-mini}
& Dart-v1
& F
& 9.55 & 9.47 & 9.54 & 9.57 & 9.57
& 9.56 & 9.53 & 9.57 & 9.56 & 9.54 \\
&
& H
& 9.95 & 9.95 & 9.97 & 9.98 & 9.97
& 9.96 & 9.97 & 9.98 & 9.98 & 9.97 \\

& Dart-qe-v1
& F
& 9.55 & 9.53 & 9.54 & 9.56 & 9.57
& 9.56 & 9.52 & 9.57 & 9.54 & 9.55 \\
&
& H
& 9.97 & 9.98 & 9.97 & 9.98 & 9.96
& 9.96 & 9.97 & 9.97 & 9.97 & 9.96 \\

& Screen-v1
& F
& 9.59 & 9.55 & 9.54 & 9.57 & 9.58
& 9.56 & 9.53 & 9.58 & 9.57 & 9.54 \\
&
& H
& 9.98 & 9.97 & 9.97 & 9.98 & 9.97
& 9.97 & 9.97 & 9.97 & 9.97 & 9.97 \\
\midrule

\multirow{6}{*}{Claude-3.7-Sonnet}
& Dart-v1
& F
& 9.73 & 9.67 & 9.65 & 9.66 & 9.67
& 9.65 & 9.64 & 9.67 & 9.66 & 9.65 \\
&
& H
& 9.98 & 9.97 & 9.99 & 9.98 & 9.99
& 9.98 & 9.99 & 9.98 & 9.98 & 9.98 \\

& Dart-qe-v1
& F
& 9.76 & 9.70 & 9.68 & 9.67 & 9.68
& 9.67 & 9.65 & 9.68 & 9.67 & 9.67 \\
&
& H
& 9.99 & 9.97 & 9.98 & 9.98 & 9.99
& 9.99 & 9.99 & 9.98 & 9.98 & 9.99 \\

& Screen-v1
& F
& 9.70 & 9.63 & 9.65 & 9.65 & 9.66
& 9.64 & 9.63 & 9.66 & 9.65 & 9.64 \\
&
& H
& 9.99 & 9.97 & 9.98 & 9.98 & 9.98
& 9.98 & 9.99 & 9.98 & 9.98 & 9.98 \\

\bottomrule
\end{tabular}
}

\caption{
Full cross-lingual evaluation results across
three screenshot datasets and ten languages.
F = Faithfulness; H = Hallucination.
}
\label{tab:multilingual_full}
\end{table*}

\paragraph{Key observations.}
Saturation persists across languages: within each model family,
Faithfulness varies by at most 0.12 and Hallucination by at most 0.07
across ten languages, far less than the between-model gaps on English.
All Hallucination scores fall within \hlsat{$[9.95, 10.00]$}, a far narrower band
than the \hlsat{3.6 to 7.0} collapse under grouped summarization, so severe
compression, not multilinguality, is the dominant production failure
mode.

\section{Failure-Mode Analysis: Full Results}
\label{app:failure_modes}

\subsection{F1: Compression Collapse Across Models and Datasets}
\label{app:compression_collapse}

Table~\ref{tab:full_collapse} reports the complete set of
(model, dataset, setting) results underlying
Figure~\ref{fig:grouped_collapse}.

\paragraph{Reading guide.}
\emph{Individual} (\texttt{ind.}) generates one unconstrained summary
per document; \emph{grouped} (\texttt{grp.}) generates a single
100-word summary per task group. The source documents are identical
across settings, isolating the effect of aggressive compression.

\begin{table*}[t]
\centering
\small
\setlength{\tabcolsep}{4pt}
\begin{tabular}{llcccccc}
\toprule
\textbf{Dataset} &
\textbf{Model} &
\textbf{Setting} &
\textbf{Faith.} &
\textbf{Compl.} &
\textbf{Underst.} &
\textbf{Hall.} &
\textbf{Avg} \\
\midrule
ECT+FindSum & \texttt{Internal-S} & ind. & 8.95 & 7.85 & 9.66 & 9.98 & 9.11 \\
 & & \cellcolor{saturatepink!15}grp. & \cellcolor{saturatepink!15}3.93 & \cellcolor{saturatepink!15}3.96 & \cellcolor{saturatepink!15}6.99 & \cellcolor{saturatepink!15}4.00 & \cellcolor{saturatepink!15}4.72 \\
\cmidrule(l){3-8}
 & \texttt{Internal-L} & ind. & 9.44 & 8.08 & 9.77 & 10.00 & 9.32 \\
 & & \cellcolor{saturatepink!15}grp. & \cellcolor{saturatepink!15}5.90 & \cellcolor{saturatepink!15}6.25 & \cellcolor{saturatepink!15}8.15 & \cellcolor{saturatepink!15}6.97 & \cellcolor{bestgreen}\textbf{6.82} \\
\cmidrule(l){3-8}
 & Claude-3.7-Sonnet & ind. & 9.08 & 8.06 & 9.83 & 10.00 & 9.24 \\
 & & \cellcolor{saturatepink!15}grp. & \cellcolor{saturatepink!15}3.75 & \cellcolor{saturatepink!15}4.05 & \cellcolor{saturatepink!15}6.99 & \cellcolor{saturatepink!15}3.63 & \cellcolor{saturatepink!15}4.60 \\
\cmidrule(l){3-8}
 & Gemini-2.5-Flash & ind. & 8.96 & 8.01 & 9.50 & 9.97 & 9.11 \\
 & & \cellcolor{saturatepink!15}grp. & \cellcolor{saturatepink!15}4.26 & \cellcolor{saturatepink!15}4.71 & \cellcolor{saturatepink!15}7.26 & \cellcolor{saturatepink!15}4.24 & \cellcolor{saturatepink!15}5.12 \\
\cmidrule(l){3-8}
 & GPT-4.1-mini & ind. & 9.44 & 8.40 & 9.76 & 10.00 & \cellcolor{bestgreen}\textbf{9.40} \\
 & & \cellcolor{saturatepink!15}grp. & \cellcolor{saturatepink!15}4.55 & \cellcolor{saturatepink!15}5.29 & \cellcolor{saturatepink!15}7.54 & \cellcolor{saturatepink!15}4.84 & \cellcolor{saturatepink!15}5.55 \\
\midrule
GovReport+BillSum & \texttt{Internal-S} & ind. & 9.07 & 7.82 & 9.62 & 10.00 & 9.13 \\
 & & \cellcolor{saturatepink!15}grp. & \cellcolor{saturatepink!15}3.48 & \cellcolor{saturatepink!15}3.19 & \cellcolor{saturatepink!15}6.48 & \cellcolor{saturatepink!15}3.83 & \cellcolor{saturatepink!15}4.24 \\
\cmidrule(l){3-8}
 & \texttt{Internal-L} & ind. & 9.42 & 8.06 & 9.68 & 10.00 & 9.29 \\
 & & \cellcolor{saturatepink!15}grp. & \cellcolor{saturatepink!15}5.86 & \cellcolor{saturatepink!15}6.01 & \cellcolor{saturatepink!15}7.90 & \cellcolor{saturatepink!15}7.00 & \cellcolor{bestgreen}\textbf{6.69} \\
\cmidrule(l){3-8}
 & Claude-3.7-Sonnet & ind. & 9.22 & 8.33 & 9.85 & 10.00 & 9.35 \\
 & & \cellcolor{saturatepink!15}grp. & \cellcolor{saturatepink!15}3.93 & \cellcolor{saturatepink!15}4.00 & \cellcolor{saturatepink!15}7.09 & \cellcolor{saturatepink!15}5.81 & \cellcolor{saturatepink!15}5.21 \\
\cmidrule(l){3-8}
 & Gemini-2.5-Flash & ind. & 9.36 & 8.38 & 9.74 & 10.00 & 9.37 \\
 & & \cellcolor{saturatepink!15}grp. & \cellcolor{saturatepink!15}4.84 & \cellcolor{saturatepink!15}4.97 & \cellcolor{saturatepink!15}7.50 & \cellcolor{saturatepink!15}4.83 & \cellcolor{saturatepink!15}5.53 \\
\cmidrule(l){3-8}
 & GPT-4.1-mini & ind. & 9.56 & 8.50 & 9.70 & 10.00 & \cellcolor{bestgreen}\textbf{9.44} \\
 & & \cellcolor{saturatepink!15}grp. & \cellcolor{saturatepink!15}4.30 & \cellcolor{saturatepink!15}4.66 & \cellcolor{saturatepink!15}7.23 & \cellcolor{saturatepink!15}4.79 & \cellcolor{saturatepink!15}5.24 \\
\bottomrule
\end{tabular}
\caption{
Complete individual-to-grouped evaluation results for F1 
(\textbf{compression collapse}). Best-performing model per 
(dataset, setting) is shown in \textbf{bold}.
}
\label{tab:full_collapse}
\end{table*}

\paragraph{Key observations.}
The full table confirms the pattern from
Figure~\ref{fig:grouped_collapse} at the per-cell level: every
model collapses under the grouped constraint, and Hallucination
falls hardest (for example, Claude-3.7-Sonnet from \hlsat{10.00 to 3.63}
on ECT+FindSum). \texttt{Internal-L} records the highest
grouped scores despite weaker individual summaries, the Goodhart
effect detailed under F4.

\paragraph{Why Hallucination collapses:
judge-compression conflation.}
The observed collapse is not primarily caused by fabricated
claims.
Manual inspection of
30
low-scoring grouped outputs reveals a more subtle failure mode.

In the grouped setting, the judge compares a
100-word summary against a reference formed by concatenating two
single-document summaries
(approximately
450 to 600
words total).
Consequently, factual details omitted for compression are often
treated as unsupported.

For example, an abstractive statement such as
``the CNMI permit program was amended''
may be penalized when the reference instead contains a detailed
legal formulation such as
``Public Law No.~110-229, enacted in 2008, amended the
U.S.-CNMI covenant.''
Although semantically consistent, the compressed version lacks
the statutory detail expected by the judge.

The result is a systematic conflation of
\emph{compression loss}
with
\emph{fabrication}.
By operating over atomic semantic units rather than surface
text overlap, the
\textbf{Semantic Scaffold}
separates these effects:
missing secondary details manifest as a
\emph{Supporting Fact Coverage gap},
rather than an artificial collapse in
Hallucination.

\subsection{F2: Ranking Instability, Full Per-Setting Breakdown}

Table~\ref{tab:rank_disagree_extended} in the main paper summarizes
the four
(dataset, setting)
configurations in which
ROUGE-L
and Direct Scoring
(DS)
select different top-performing models.
Here, we provide the complete per-setting breakdown underlying
that disagreement.

\paragraph{Why the rankings diverge.}
In the individual setting Claude-3.7-Sonnet tops ROUGE-L: its more
extractive style overlaps more with the references. GPT-4.1-mini tops
Direct Scoring with more abstractive summaries the judge rates higher on
faithfulness and coherence. In the grouped setting \texttt{Internal-L}
leads Direct Scoring but not ROUGE-L, the same F4 non-compliance effect.
Because the disagreement holds in every configuration, metric choice
alone, not noise, decides which system ships.

\subsection{Scaffold Metrics Under Grouped Summarization}
\label{app:scaffold_grouped}

Figure~\ref{fig:scaffold_spread} shows that scaffold-based metrics
retain meaningful separation even when holistic Direct Scoring
collapses under grouped summarization.
Table~\ref{tab:scaffold_grouped} provides the complete metric
breakdown averaged across both grouped benchmark datasets.

\begin{table*}[t]
\centering
\footnotesize
\setlength{\tabcolsep}{4pt}

\resizebox{\textwidth}{!}{%
\begin{tabular}{llcccccccc}
\toprule
\textbf{Rank} &
\textbf{Model} &
\textbf{F.\ Faith.} &
\textbf{F.\ Cov.} &
\textbf{F.\ Focus} &
\textbf{Main FC} &
\textbf{Supp.\ FC} &
\textbf{FPS} &
\textbf{QPS} &
\textbf{EPS} \\
\midrule

1 & Claude-3.7-Sonnet
& \cellcolor{bestgreen}0.986 & \cellcolor{bestgreen}0.539 & 0.752
& \cellcolor{bestgreen}0.693 & \cellcolor{bestgreen}0.316 & 0.453 & 0.615 & \cellcolor{bestgreen}0.801 \\

2 & Gemini-2.5-Flash
& 0.961 & 0.493 & 0.866
& 0.674 & 0.260 & \cellcolor{bestgreen}0.481 & 0.655 & 0.789 \\

3 & GPT-4.1-mini
& 0.975 & 0.444 & 0.835
& 0.627 & 0.216 & 0.476 & \cellcolor{bestgreen}0.693 & 0.773 \\

4 & \texttt{Internal-S}
& 0.807 & 0.287 & 0.992
& 0.400 & 0.103 & 0.353 & 0.628 & 0.596 \\

5 & \texttt{Internal-L}
& 0.726 & 0.288 & 1.000
& 0.385 & 0.092 & 0.343 & 0.649 & 0.598 \\

\midrule

\multicolumn{2}{l}{\cellcolor{scaffoldblue!25}\textbf{Best-to-worst spread}}
& \cellcolor{scaffoldblue!25}0.260 & \cellcolor{scaffoldblue!25}0.252 & \cellcolor{scaffoldblue!25}0.248
& \cellcolor{scaffoldblue!25}0.308 & \cellcolor{scaffoldblue!25}0.224 & \cellcolor{scaffoldblue!25}0.138 & \cellcolor{scaffoldblue!25}0.078 & \cellcolor{scaffoldblue!25}0.205 \\

\multicolumn{2}{l}{\cellcolor{scaffoldblue!25}\textbf{Relative spread}}
& \cellcolor{scaffoldblue!25}1.4$\times$
& \cellcolor{scaffoldblue!25}1.9$\times$
& \cellcolor{scaffoldblue!25}1.3$\times$
& \cellcolor{scaffoldblue!25}1.8$\times$
& \cellcolor{bestgreen}\textbf{3.4$\times$}
& \cellcolor{scaffoldblue!25}1.4$\times$
& \cellcolor{scaffoldblue!25}1.1$\times$
& \cellcolor{scaffoldblue!25}1.3$\times$ \\

\bottomrule
\end{tabular}
}

\caption{
Scaffold-based metrics averaged across grouped
multi-document benchmark settings.
}
\label{tab:scaffold_grouped}
\end{table*}

\paragraph{Column interpretation.}
Faithfulness (\textbf{F.\ Faith.}) stays high, so the grouped drop comes
from omission, not fabrication, while overall fact coverage
(\textbf{F.\ Cov.}) falls sharply under the 100-word cap. \textbf{Main
Fact Coverage} spreads more widely (\hlbest{$1.8\times$}) than total coverage, so
stronger models keep the \emph{right} facts rather than simply more of
them, and \textbf{Supporting Fact Coverage} spreads widest (\hlbest{$3.4\times$}):
supporting detail is the first casualty of compression. \textbf{FPS}
combines main-point recall with a supporting-detail penalty, and
\textbf{QPS} and \textbf{EPS} stay comparatively stable while still separating models.

Compared with the proprietary results, the ordering of spreads inverts here: Main and Supporting Fact Coverage separate models most ($1.8\times$ and $3.4\times$), while FPS, QPS, and EPS compress (0.138, 0.078, 0.205 best-to-worst). This is expected under a uniform 100-word cap, which pushes every system toward the same operating point on the main/supporting trade-off that FPS and QPS are built to expose. On this data the hierarchy-aware components are the informative axes and the aggregates are best read as summaries of them.

\paragraph{A revealing failure signature.}
\texttt{Internal-L} reaches perfect F.\ Focus (\hlnum{1.000}), meaning
every sentence matches some scaffold fact, yet records the lowest
Supporting Fact Coverage (\hlsat{0.092}): it stays on-topic by filtering to a
narrow slice of content instead of synthesizing across documents.

\subsection{F4: Prompt Relaxation Does Not Recover Performance}
\label{app:prompt_relaxation}

A natural response to the compression collapse observed in
F1 is to relax the summarization constraint itself.
If the
100-word cap
is driving the degradation, then allowing longer and more
flexible summaries should restore evaluation quality.

To test this hypothesis, we compare two grouped summarization
prompts:

\paragraph{Prompt v1 (constrained summarization).}
\emph{
``Generate a concise summary (less than 100 words) that
captures the main topic, purpose, and key details of the
documents\ldots''
}

\paragraph{Prompt v2 (production-oriented summarization).}
\emph{
``Generate a summary of these documents as it would appear
attached to a record in a service management platform.
Focus on the information most useful for a service agent or
knowledge worker to understand the document's purpose, key
findings, and any actionable outcomes without reading the full
document.
Use clear, plain language proportional to the document's
complexity.''
}

Table~\ref{tab:prompt_v1v2} compares the two settings.
Prompt
v1
uses the original grouped
100-word cap,
while
v2
removes the length constraint and instead relies on a
production-style summarization objective.

\begin{table}[!h]
\centering
\small
\setlength{\tabcolsep}{4pt}

\begin{tabular}{p{4cm}cccc}
\toprule
\textbf{Dataset} &
\textbf{RG-L v1} &
\textbf{RG-L v2} &
\textbf{DS v1} &
\textbf{DS v2} \\
&
\textbf{(v1)} &
\textbf{(v2)} &
\textbf{(v1)} &
\textbf{(v2)} \\
\midrule

Dart-v1
& 0.358 & \cellcolor{saturatepink!20}0.275 & 8.99 & \cellcolor{saturatepink!20}8.86 \\

Dart-qe-v1
& 0.355 & \cellcolor{saturatepink!20}0.251 & 8.84 & \cellcolor{saturatepink!20}8.77 \\

\makecell[l]{GovReport+\\BillSum}
& 0.228 & \cellcolor{saturatepink!20}0.102 & 6.69 & \cellcolor{saturatepink!20}5.80 \\

ECT+FindSum
& 0.111 & \cellcolor{saturatepink!20}0.090 & 6.82 & \cellcolor{saturatepink!20}5.76 \\

\bottomrule
\end{tabular}

\caption{
Prompt comparison:
v1
(100-word cap)
versus
v2
(no explicit cap; production-oriented instruction),
averaged across five models.
}
\label{tab:prompt_v1v2}
\end{table}

\paragraph{Key observations.}
The grouped Direct Score leader is lower under v2 on all four datasets
(for example, 6.69 to 5.80 on GovReport+BillSum and 6.82 to 5.76 on
ECT+FindSum), and the individual ROUGE-L leader also decreases. Because v2
changes both framing and the length instruction, these results do not
isolate either factor or show that prompt wording cannot recover performance.

\section{Metric Design Rationale and Edge Cases}
\label{app:metric_design}

\subsection{Why a Multiplicative Form for FPS and QPS}

FPS and QPS multiply a main-point recall term ($M_s/M_c$) by a
supporting-detail penalty ($1-S_s/S_c$) rather than adding them.
The product imposes a \emph{joint} constraint: a summary scores
well only if it both preserves central content \emph{and}
compresses secondary detail. An additive form would not do
this: a weighted sum can still reward a summary that copies
every supporting fact, or under-penalize one that drops core
content. With the product, failure on either axis pulls the
score toward zero: no main-fact coverage gives FPS\,=\,0
regardless of supporting detail, and reproducing all supporting
facts gives FPS\,=\,0 regardless of recall. This encodes the
basic assumption of summarization: keep the priority structure,
compress the rest.

FPS is most informative for abstractive, length-constrained
tasks where the goal is to preserve a document's informational
hierarchy: bug-report digests, executive briefings, legal case
summaries, news. It is a poor standalone target for regulated or
high-recall workflows (clinical trial protocols, regulatory
filings, audit reports) where retaining supporting detail is
desirable; there, read FPS alongside FC and Supporting Fact
Coverage.

\subsection{EPS: Continuous Rather Than Binary Entity Evaluation}

FineSurE flags entity consistency with a binary, sentence-level
criterion: a sentence either has an entity error or it does not.
That misses the gradation in practice, where a summary may name
an entity correctly yet drop, compress, or distort some of its
attributes. EPS instead works at the attribute level, measuring
the fraction of scaffold-defined entity attributes preserved:

\begin{itemize}\small

\item
A summary that mentions
``Windows Hello''
but omits its independence from the
privacy-settings toggle preserves only
$1/2$
attributes for that entity.

\item
A summary that mentions an entity name but misattributes all
associated properties contributes
0,
equivalent to omitting the entity entirely.

\item
A summary preserving all entity attributes across the scaffold
receives
EPS\,=\,1.0.

\end{itemize}

This granularity matters in legal, technical, and enterprise
settings, where an entity carries many important attributes
(contract clauses, error codes, system components, legal
references) and partial preservation differs meaningfully from
omission.

\subsection{Reporting hierarchy without an inverse penalty}

Main and supporting coverage can be reported separately without assuming
that supporting content should reduce quality. Designing a validated
hierarchy-aware aggregate requires held-out evaluation and remains future
work.

\section{Failure-Mode Figures: Extended Captions and Guided Reading}
\label{app:ffigs}

\begin{figure*}[t]
\centering
\includegraphics[width=0.85\textwidth]
{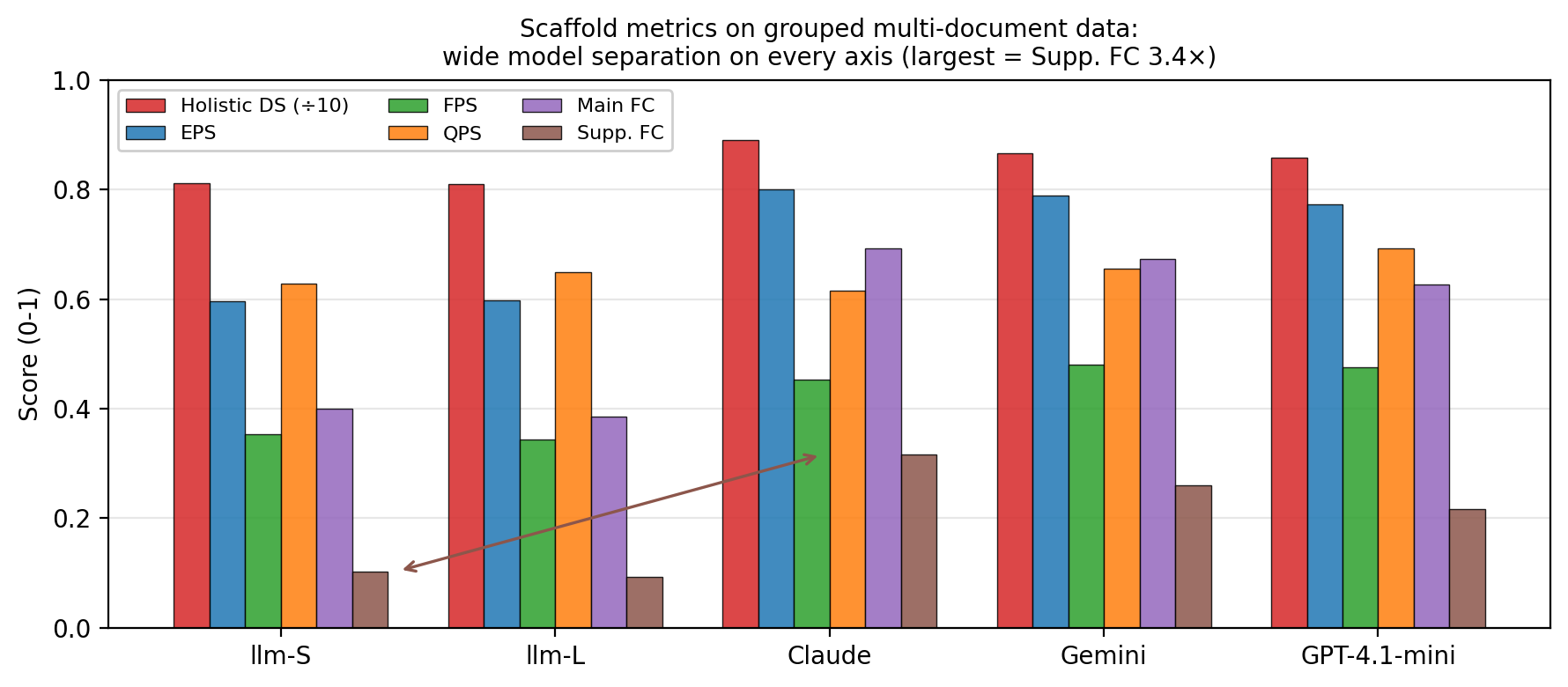}
\caption{Scaffold metrics on grouped multi-document summarization
(five models). Each axis keeps meaningful spread (Supporting Fact
Coverage $3.4\times$, Main Fact Coverage $1.8\times$, EPS
$1.3\times$, FPS $1.4\times$) while holistic Direct Scoring
(\textit{red}, shown as DS$/10$) compresses into 0.81 to 0.89. The
widest spread, Supporting Fact Coverage, matches the F1 diagnosis:
under compression, supporting details fail first.}
\label{fig:scaffold_spread}
\end{figure*}

Figure~\ref{fig:scaffold_spread} plots the scaffold metrics on the
same grouped data as Figure~\ref{fig:grouped_collapse} in the main
text.

In Figure~\ref{fig:grouped_collapse}, Direct Scoring and Hallucination both
decrease in the grouped setting. Internal-L degrades least and also
under-complies with the 100-word limit, but these data do not establish that
non-compliance causes the difference. Figure~\ref{fig:scaffold_spread}
localizes coverage differences without identifying their cause.

\section{Complete Prompt Templates with Design Notes}
\label{app:prompts}

All scaffold extraction and evaluation prompts are provided verbatim.
Each prompt is prefixed with a design note explaining the key choices.

\noindent\textbf{Common settings for all prompts:}
model = GPT-4.1, \texttt{temperature\,=\,0.1},
\texttt{max\_tokens\,=\,4096},
\texttt{response\_format: json\_object}.
The low temperature (0.1 rather than 0.0) prevents perfectly
deterministic but occasionally degenerate JSON while keeping
scores highly reproducible.

\subsection{Fact Extraction}

\textit{Design note.}
The 20--100 range spans short documents (screenshots yield 8--15 facts)
and long ones (contracts yield 60--100). The three-tier label routes
metadata and boilerplate to ``other'' so it cannot affect scores, and
ordering main points first doubles as a consistency check on the
extraction.

\begin{promptbox}[Fact Extraction Prompt]
\begin{lstlisting}
System: You are a precise information extraction assistant.
        Return valid JSON only. No preamble, no explanation.

User:
Extract 20-100 key facts from the document below.

Each fact must be:
  - Concise (at most 2-3 entities per fact)
  - Self-contained (understandable without the full document)
  - Labeled with exactly one of three types:
      "main point"        -- central claim, finding, or action
      "supporting detail" -- context, evidence, elaboration
      "other"             -- metadata, dates, boilerplate

Order: all main points first, then supporting details, then other.
Do not include duplicate or redundant facts.

Output JSON format:
{
  "key_facts": [
    {"id": "1", "fact": "...", "type": "main point"},
    {"id": "2", "fact": "...", "type": "supporting detail"},
    ...
  ]
}

Document:
{{DOCUMENT_TEXT}}
\end{lstlisting}
\end{promptbox}

\subsection{Question Extraction}

\textit{Design note.}
Questions capture the document's implied agenda. The expected answer
and source-sentence provenance make the scaffold auditable and let the
question-alignment step (\S\ref{app:qalign}) align at the sentence
level without re-reading the full document.

\begin{promptbox}[Question Extraction Prompt]
\begin{lstlisting}
System: You are a precise information extraction assistant.
        Return valid JSON only. No preamble, no explanation.

User:
Extract 20-100 key questions from the document below.

For each question provide:
  - A brief expected answer (1-2 sentences, from the document)
  - A type: "main point", "supporting detail", or "other"
  - The source (e.g. "Sentence 3" or "Paragraph 2")

Order: main-point questions first, then supporting, then other.

Output JSON format:
{
  "questions": [
    {
      "id": "1",
      "question": "...",
      "answer": "...",
      "type": "main point",
      "source": "Sentence 3"
    },
    ...
  ]
}

Document:
{{DOCUMENT_TEXT}}
\end{lstlisting}
\end{promptbox}

\subsection{Entity Attributes Extraction}

\textit{Design note.}
Entity extraction is deliberately broad (people, organizations,
products, legal instruments, technical and abstract concepts), and
each attribute is a single statement so the entity-preservation prompt
(\S\ref{app:entpres}) can check it independently.

\begin{promptbox}[Entity Extraction Prompt]
\begin{lstlisting}
System: You are a precise information extraction assistant.
        Return valid JSON only. No preamble, no explanation.

User:
Extract all key entities from the document below.

Entities include: people, organisations, products, concepts,
legal instruments, technical terms, or any named object central
to the document.

For each entity list its key attributes. Each attribute must:
  - Be a single concise factual statement (one claim)
  - Be labeled: "main point", "supporting detail", or "other"

Output JSON format:
{
  "entities": [
    {
      "entity_name": "...",
      "entity_type": "person | org | product | concept | other",
      "attributes": [
        {"attribute": "...", "type": "main point"},
        {"attribute": "...", "type": "supporting detail"}
      ]
    }
  ]
}

Document:
{{DOCUMENT_TEXT}}
\end{lstlisting}
\end{promptbox}

\subsection{Key Fact Alignment Evaluation}
\label{app:falign}

\textit{Design note.}
This prompt computes $M_s$ and $S_s$ in the FPS formula
(Equation~1 in the main paper) by determining which scaffold facts
are inferable from the summary.
The ``sentence IDs'' output enables item-level audit: a practitioner
can inspect exactly which summary sentences support each scaffold
fact, identifying precisely which main-point facts were missed.
The justification field discourages shallow keyword matching.

\begin{promptbox}[Fact Alignment Prompt]
\begin{lstlisting}
System: You are a precise evaluation assistant.
        Return valid JSON only. No preamble, no explanation.

User:
You receive a document summary and key facts from the original
document.

For each key fact, determine if it can be inferred from the
summary. "Inferred" means the fact's meaning is clearly conveyed,
even in different words. Do NOT mark a fact as inferred if
the summary is only tangentially related.

For inferred facts: list the summary sentence ID(s) that support
it, and explain why in 1 sentence.

Include all facts in the output, even those not inferred.

Output JSON format:
{
  "fact_alignment": [
    {
      "fact_id": "1",
      "inferred": true,
      "sentence_ids": [2],
      "justification": "Sentence 2 states that..."
    },
    {
      "fact_id": "3",
      "inferred": false,
      "sentence_ids": [],
      "justification": "No sentence in the summary covers this."
    }
  ]
}

Key facts (with types):
{{FACTS_JSON}}

Summary (sentences numbered starting at 1):
{{NUMBERED_SUMMARY}}
\end{lstlisting}
\end{promptbox}

\subsection{Key Question Alignment Evaluation}
\label{app:qalign}

\textit{Design note.}
Structurally analogous to \S\ref{app:falign} but operating on questions.
A question is ``answered'' if the summary provides enough information
for a reader to answer it, not if the question is literally asked
and answered. This softer criterion is why QPS tends to exceed FPS across models.

\begin{promptbox}[Question Alignment Prompt]
\begin{lstlisting}
System: You are a precise evaluation assistant.
        Return valid JSON only. No preamble, no explanation.

User:
You receive a document summary and key questions about the
original document.

For each question, determine if the summary provides enough
information for a reader to answer it. If yes, list the
sentence ID(s) containing the relevant information and
explain in 1 sentence.

Output JSON format:
{
  "question_alignment": [
    {
      "question_id": "1",
      "answered": true,
      "sentence_ids": [1, 3],
      "justification": "..."
    },
    {
      "question_id": "4",
      "answered": false,
      "sentence_ids": [],
      "justification": "The summary does not address..."
    }
  ]
}

Key questions (with types):
{{QUESTIONS_JSON}}

Summary (sentences numbered starting at 1):
{{NUMBERED_SUMMARY}}
\end{lstlisting}
\end{promptbox}

\subsection{Entity Preservation Evaluation}
\label{app:entpres}

\textit{Design note.}
Checks attribute-by-attribute preservation rather than entity-level
presence, enabling the continuous EPS score described in
Appendix~\ref{app:metric_design}.
Only attributes that appear in the summary are returned; the EPS
denominator is always the total scaffold attribute count regardless
of what the summary mentions.

\begin{promptbox}[Entity Preservation Prompt]
\begin{lstlisting}
System: You are a precise evaluation assistant.
        Return valid JSON only. No preamble, no explanation.

User:
You receive a document summary and a list of key entities with
their attributes, extracted from the original document.

For each entity that appears in the summary, list the attributes
that are correctly and faithfully represented. An attribute is
"correctly represented" if its meaning is preserved in the
summary, even in different words.

Do NOT include entities absent from the summary.
Do NOT include attributes that are distorted or only partially
correct.
Do NOT invent attributes not listed in the input.

Output JSON format:
{
  "entity_preservation": [
    {
      "entity_name": "...",
      "preserved_attributes": [
        "Exact text of the preserved attribute from the input list"
      ]
    }
  ]
}

Entities and attributes from the scaffold:
{{ENTITIES_JSON}}

Summary to evaluate:
{{SUMMARY_TEXT}}
\end{lstlisting}
\end{promptbox}

\subsection{Direct Score Evaluation (holistic baseline)}
\label{app:direct_score}

\textit{Design note.}
This is the holistic LLM-as-judge prompt used for all direct-scoring
baselines.
The 1 to 10 scale with explicit anchors at 1, 5, and 10 matches prior
work (SummEval, FineSurE) and provides meaningful gradations.
Hallucination is framed as its \emph{absence} (10 = no hallucination)
so higher is always better across all four dimensions.
The judge receives only the raw source document, not the scaffold,
making it a true holistic baseline without hierarchy information.
The \texttt{rationale} field aids human auditing of low-scoring cells.

\begin{promptbox}[Direct Score (Holistic Baseline) Prompt]
\begin{lstlisting}
System: You are a rigorous and impartial evaluation assistant.
        Return valid JSON only. No preamble, no explanation.

User:
Evaluate the quality of the summary against the original
document. Rate on four dimensions using a 1-10 scale where
10 is perfect.

Dimension definitions:
1. Faithfulness
   10 = perfectly faithful; no distortions or omissions of
        information that is present
    5 = mostly faithful; minor inaccuracies
    1 = fabricates or contradicts the document

2. Completeness
   10 = covers all key information needed for understanding
    5 = covers some key points but misses important ones
    1 = misses most key information

3. Understandability
   10 = perfectly clear, logical, easy to follow
    5 = somewhat clear; has clarity or organisation issues
    1 = incomprehensible or severely disorganised

4. Hallucination (absence of hallucination)
   10 = no hallucination; all claims supported by the source
    5 = some claims not clearly supported; questionable
    1 = severe fabrications; significant unsupported claims

Output JSON format:
{
  "faithfulness": N,
  "completeness": N,
  "understandability": N,
  "hallucination": N,
  "rationale": "1-2 sentence explanation of the scores"
}

Original document:
{{DOCUMENT_TEXT}}

Summary to evaluate:
{{SUMMARY_TEXT}}
\end{lstlisting}
\end{promptbox}

\end{document}